\documentclass[10pt,twocolumn,letterpaper]{article}

\usepackage[pagenumbers]{iccv} 

\definecolor{iccvblue}{rgb}{0.21,0.49,0.74}
\usepackage[pagebackref,breaklinks,colorlinks,allcolors=iccvblue]{hyperref}

\usepackage{array}
\usepackage{xcolor}
\usepackage{graphicx}
\usepackage{amsmath}
\usepackage{amssymb}
\usepackage{booktabs}
\usepackage{float}
\usepackage{comment}
\usepackage[accsupp]{axessibility}
\usepackage{tikz}

\usepackage{tabularx}
\usepackage{multirow}
\usepackage{adjustbox}
\usepackage{makecell}
\usepackage{placeins}
\usepackage{environ}
\usepackage[normalem]{ulem}
\usepackage{pifont}
\usepackage{subcaption} 
\usepackage{wrapfig}
\usepackage{url}
\usepackage{epsfig}
\usepackage{enumitem}
\usepackage{multibib}
\usepackage{numprint}
\usepackage[figuresleft]{rotating}
\usepackage{caption}
\usepackage{bbding}

\usepackage[utf8]{inputenc} 
\usepackage[T1]{fontenc}    
\usepackage{amsfonts}       
\usepackage{nicefrac}       
\usepackage{microtype}      

\usepackage{algorithm}
\usepackage{algorithmicx}
\usepackage{algpseudocodex}
\usepackage{varwidth}
\usepackage{colortbl}
\usepackage[normalem]{ulem}

\definecolor{c1}{HTML}{765D97} 
\definecolor{c2}{HTML}{7BA67D}
\definecolor{c3}{HTML}{fc6160}
\definecolor{myblue}{HTML}{E6F3FC} 
\definecolor{mygray}{HTML}{DBE2E9} 
\definecolor{mygreen}{HTML}{006400} 
\usepackage{tikz}

\newcommand{\E}{\mathbb{E}}

\newcommand{\rd}{{\mathrm{d}}}

\usepackage{xspace}

\makeatletter
\DeclareRobustCommand\onedot{\futurelet\@let@token\@onedot}
\def\@onedot{\ifx\@let@token.\else.\null\fi\xspace}

\def\eg{\emph{e.g}\onedot} 
\def\ie{\emph{i.e}\onedot}

\makeatother

\title{SCFlow: Implicitly Learning Style and Content Disentanglement with Flow Models}

\author{Pingchuan Ma* \hspace{5mm} Xiaopei Yang* \hspace{5mm} Yusong Li \hspace{5mm} \\ Ming Gui \hspace{5mm} Felix Krause \hspace{5mm} Johannes Schusterbauer \hspace{5mm} Bj\"orn Ommer\\ \\
CompVis @ LMU Munich \hspace{1cm} Munich Center for Machine Learning (MCML)\\
}

\begin{document}
\twocolumn[{%
\renewcommand\twocolumn[1][]{#1}%
\maketitle
\vspace{-10mm}
\begin{center}
    \centering
    \includegraphics[width=.75\linewidth]{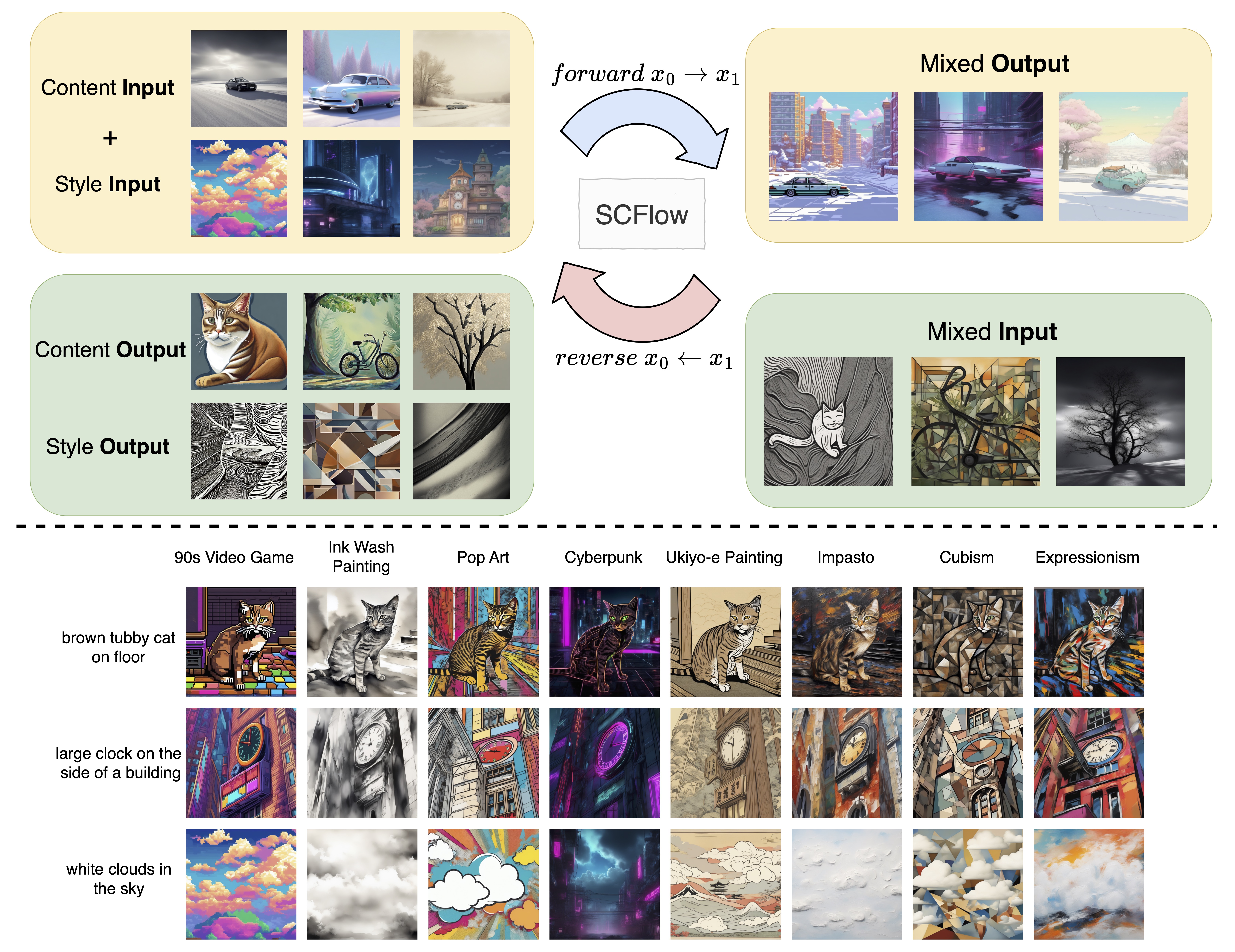}
    \captionof{figure}{\textit{Top:} The proposed \textit{SCFlow} works bidirectionally, enabling style-content merging (\textit{forward}) and disentangling (\textit{reverse}) with a single model. \textit{Bottom:} Our curated dataset to facilitate training. For details, see \Cref{sec:method}.}
    \label{fig:teaser}
\end{center}
}]
{
\def\thefootnote{*}\footnotetext{Equal Contribution}

\begin{abstract}
Explicitly disentangling style and content in vision models remains challenging due to their semantic overlap and the subjectivity of human perception. Existing methods propose separation through generative or discriminative objectives, but they still face the inherent ambiguity of disentangling intertwined concepts. Instead, we ask: \textit{Can we bypass explicit disentanglement by learning to merge style and content invertibly, allowing separation to emerge naturally?}

We propose SCFlow, a flow-matching framework that learns bidirectional mappings between entangled and disentangled representations. Our approach is built upon three key insights: 1) Training solely to merge style and content, a well-defined task, enables invertible disentanglement without explicit supervision; 2) flow matching bridges on arbitrary distributions, avoiding the restrictive Gaussian priors of diffusion models and normalizing flows; and 3) a synthetic dataset of $510{,}000$ samples ($51$ styles $\times 10{,}000$ content samples) was curated to simulate disentanglement through systematic style-content pairing.

Beyond controllable generation tasks, we demonstrate that SCFlow generalizes to ImageNet-1k and WikiArt in zero-shot settings and achieves competitive performance, highlighting that disentanglement naturally emerges from the invertible merging process. Code and dataset:  \url{https://github.com/CompVis/SCFlow}
\end{abstract}    
\section{Introduction}
\label{sec:intro}

Effectively disentangling style and content in computer vision remains challenging due to their semantic overlap and the subjectivity of human perception. 
While existing models learn diverse latent representations for these attributes, defining explicit boundaries between them is an open problem.
This problem in our community spans two key paradigms: generative approaches, that try to manipulate style and content, \eg, via style transfer~\cite{Gatys2016,johnson2016perceptual,kotovenko2021rethinking,wang2024instantstyle}, image editing~\cite{Qi_2024,hertz2024style,shah2024ziplora,frenkel2024implicit,stracke2024ctrloralter}, and discriminative ones that seek effective representations~\cite{somepalli2024CSD,wang2023evaluating} through contrastive learning~\cite{sanakoyeu2021improving,radford2021learning,dino} or classification task~\cite{saleh2015large,deng2009imagenet}. 

While these approaches have achieved impressive results, they both rely on defining explicit separation criteria for inherently ambiguous concepts, which reside in the subjective nature of human perception.
Recent advances refine the generative paradigm through multi-modal conditioning, including edge maps~\cite{zhang2023controlNet}, text prompts~\cite{ldm,gal2022image}, and CLIP~\cite{radford2021learning} embeddings~\cite {ramesh2022hierarchical}. On the other hand, hierarchical analysis~\cite{voynov2023p,stracke2024cleandift,zhang2023tale,tang2023emergent} of intermediate features by generative models~\cite{ldm,ramesh2021zero_dalle1,song2020denoising_ddim,ho2020denoising} reveal that distinct layers capture attributes like shape and color, suggesting implicit disentanglement cues in their architecture. However, these approaches still operate within the separation perspective, which inherently struggles with the ambiguity of defining where the boundary lies between style and content.

This ambiguity poses a fundamental limitation; direct supervision for disentanglement is less feasible due to the lack of clean ``ground-truth'' style/content pairs and the subjective nature of their definitions.
This limitation raises an intriguing question: \textbf{Instead of tackling disentanglement directly, can we circumvent its challenges by learning to merge style and content in an invertible manner?}
Merging style and content is comparatively more straightforward with clear and well-defined data, as demonstrated by prior work on style transfer and image editing~\cite{Qi_2024,wang2024instantstyle,zhang2023controlNet}.
Disentanglement can emerge naturally by reversing the blend if the merging process is \textit{invertible}.
Motivated by this, we propose \textbf{SCFlow} to implicitly learn the disentanglement. Rather than enforcing explicit separation as the learning objective, SCFlow learns a bidirectional function to merge style and content within a semantically structured latent space (compared to the pixel space), developing disentangled representations as an emergent property of invertibility, without reliance on pixel-space or spatial biases.

To recover individual components from the merged representation, we treat content, style, and their mixture as distinct data distributions. However, a key challenge arises: to recover the original style and content from the merged output, \ie, mapping from the blended distribution to the disentangled one, requires the learned merging process to be invertible. While many modern generative models, such as normalizing flows~\cite{dinh2014nice,dinh2016density,kingma2018glow} and diffusion models~\cite{ho2020denoising,song2021scorebased_sde,karras2022elucidating,fuest2024diffusion}, offer invertibility, they commonly rely on a restrictive assumption that one end of the mapping follows a standard Gaussian distribution, limiting their suitability for bidirectional mapping of style and content. 
To address this, SCFlow employs flow matching (FM)~\cite{lipman2022flow,dao2023flow_lfm,albergo2023stochastic,albergo2022building}, which learns continuous bidirectional mappings between arbitrary distributions without stochastic diffusion steps. Unlike diffusion models that require noise-based transitions~\cite{song2020denoising_ddim,song2021scorebased_sde,ldm}, FM directly maps between blended and disentangled data distributions. 
By training solely on the merging process from content/style pairs to their entangled mixture, 
SCFlow implicitly learns to invert the blend, isolating style and content features to satisfy the invertibility. Thus, disentanglement arises not from explicit supervision but from the invertibility.

As aforementioned, we cannot define clear boundaries between style and content, nor do we have access to meaningful disentangled representations. While SCFlow assumes access to disentangled style/content pairs and their blended counterparts, real-world datasets for these attributes~\cite{saleh2015large,wilber2017bam,somepalli2024CSD,deng2009imagenet,laion-5b} rarely provide such aligned examples. 
Following the similar principle that disentanglement is hard but blending is tractable,  we leverage the vast amount of successful works~\cite{Qi_2024,wang2024instantstyle,zhang2023controlNet} tackling style transfer and image editing to simulate disentanglement. By generating stylized images that systematically pair specific content and style, we curated a dataset of $510{,}000$ samples, spanning $51$ artistic styles and $10{,}000$ content instances, where every style is applied to every content with full combinatorial coverage. 
Unlike previous datasets, our design enables SCFlow to learn disentanglement by observing how style and content vary independently, obtaining style invariance under changing content, and vice versa. 
Therefore, SCFlow implicitly infers disentangled style and content by learning how to merge them, even though ``clean'' representations are never directly observed. 
In summary, our core contributions are: 

\begin{itemize}  
    \item \textbf{SCFlow}: We present a framework that learns disentanglement implicitly by invertibly merging style and content with flow matching (\cref{sec:method}), sidestepping the difficulties imposed by explicit separation.
    \item \textbf{Combinatorial Dataset}: We curated a large-scale dataset (\cref{sec:data_construction}) consisting of $51$ styles $\times$ $10{,}000$ content pairs. It offers full combinatorial coverage, addressing the lack of aligned data in existing datasets for style and content for further systematic analyses.
    \item \textbf{Generalizable Disentanglement}: SCFlow learns pure style and content representations that not only enable blending and disentangling on our stylized dataset (\cref{sec:experiemtns_our_data}), but also generalize to unseen data. It achieves competitive performance in style retrieval on WikiArt~\cite{saleh2015large} and content recognition on ImageNet~\cite{deng2009imagenet}, demonstrating the transferability of the features (\cref{sec:imagenet_wikiart}).
\end{itemize}  
\section{Related Works}
\label{sec:related}

\subsection{Diffusion and Flow  Models}

\label{sec:rela:FM}
Diffusion models and Flow Matching represent two prominent paradigms in generative modeling. Diffusion models, as introduced by \citet{sohl2015deep} and further advanced by \citet{ho2020denoising, song2020denoising_ddim}, rely on a \textit{forward diffusion process} that incrementally adds Gaussian noise to data until the distribution converges to an isotropic Gaussian prior. A corresponding \textit{reverse diffusion process} is then learned to denoise and recover the original data distribution. Notably, inversion techniques—such as denoising diffusion implicit models (DDIM) inversion \cite{song2020denoising_ddim} allow the network to add noise to the image, which can be reversed to recover the original sample. Together with other SDE-based approaches~\cite{meng2021sdedit, hertz2022prompt}, these inversion techniques are particularly valuable in image-editing applications where controlled modifications are required.

In contrast, Flow Matching methods~\cite{lipman2022flow, rectifiedflow_iclr23, neklyudov2023action} are not dependent on an isotropic Gaussian prior and instead allow the use of arbitrary source distributions~\cite{liu2025flowing, gui2024depthfm, 2023boosting}. Flow Matching can interpolate between any possibly structured data distributions by learning an optimal transport conditional probability path via Ordinary Differential Equations (ODEs)~\cite{chen2018neural}. For instance, \citet{2023boosting} trained a flow model to map between low- and high-resolution image representations, while \citet{gui2024depthfm} established a mapping between images and depth maps. The learned mappings can, in some cases, be unidirectional due to conditioning constraints. In contrast, the works by \citet{liu2025flowing} and \citet{he2025flowtok} demonstrate a mapping between text and images by directly training the flow model without extra conditions, enabling the bidirectional mapping.

\subsection{Style and Content Representation}
Research on the style and content of images has mainly followed two directions, one of which is discriminative tasks. ~\citet{karayev2013recognizing} and ~\citet{saleh2015large} are oriented toward classification and similarity metrics in visual style. Works from ~\citet{somepalli2024measuring,wang2023evaluating} train style descriptors based on their semantic style using contrastive learning with synthetic or curated data. Another line of works~\cite{he2019momentum,radford2021learning,dino,sanakoyeu2021improving} focuses on extracting robust content descriptors, while given the data or training procedure, they can inevitably contain irrelevant information.

On the other hand, a growing amount of research has shifted toward generative tasks. The pioneering works by~\cite{gatys2016image} marked the beginning of the style transfer era, which defines style as Gram Matrices of VGG~\cite{simonyan2015deepconvolutionalnetworkslargescale} network features. 
Various methods have been proposed to improve different aspects of style representation, ranging from efficiency-focused approaches like~\cite{johnson2016perceptual,li2019learning,wang2023microast} to quality-centric ones such as~\cite{wang2020glstylenet,kotovenko2021rethinking,zuo2022style,zhang2022domain}. 
More recently, Text-driven synthesis models enable stylized editing, as presented in \citet{hertz2022prompt} and \citet{gal2022image}. Furthermore, \cite{li2023blipdiffusion,Qi_2024,xing2024csgo,frenkel2024implicit} introduced methods to extract specific features from reference images for controllable generation. Rather than merely replicating style features, these approaches aim to provide more refined control over the generated content. For instance, B-LoRA~\cite{frenkel2024implicit} implicitly separates style and content from a reference image using low-rank adaptation~\cite{hu2021lora}. However, this does not produce explicit representations of these attributes. DEADiff~\cite{Qi_2024} and CSGO~\cite{xing2024csgo} propose to inject explicit features extracted from content/style reference images into the diffusion model to achieve image-driven style transfer. Yet, they do not analyze the semantic meaning of the extracted features.~\citet{gandikota2025sliderspace} trains LoRA adapters on principal components of CLIP embeddings from generated images to ensure semantic orthogonality. However, their approach requires retraining on different data manifolds while requiring extra semantic labeling.

\section{SCFlow}
\label{sec:method}

\begin{figure}[hb]
    \centering
    \includegraphics[width=1\linewidth]{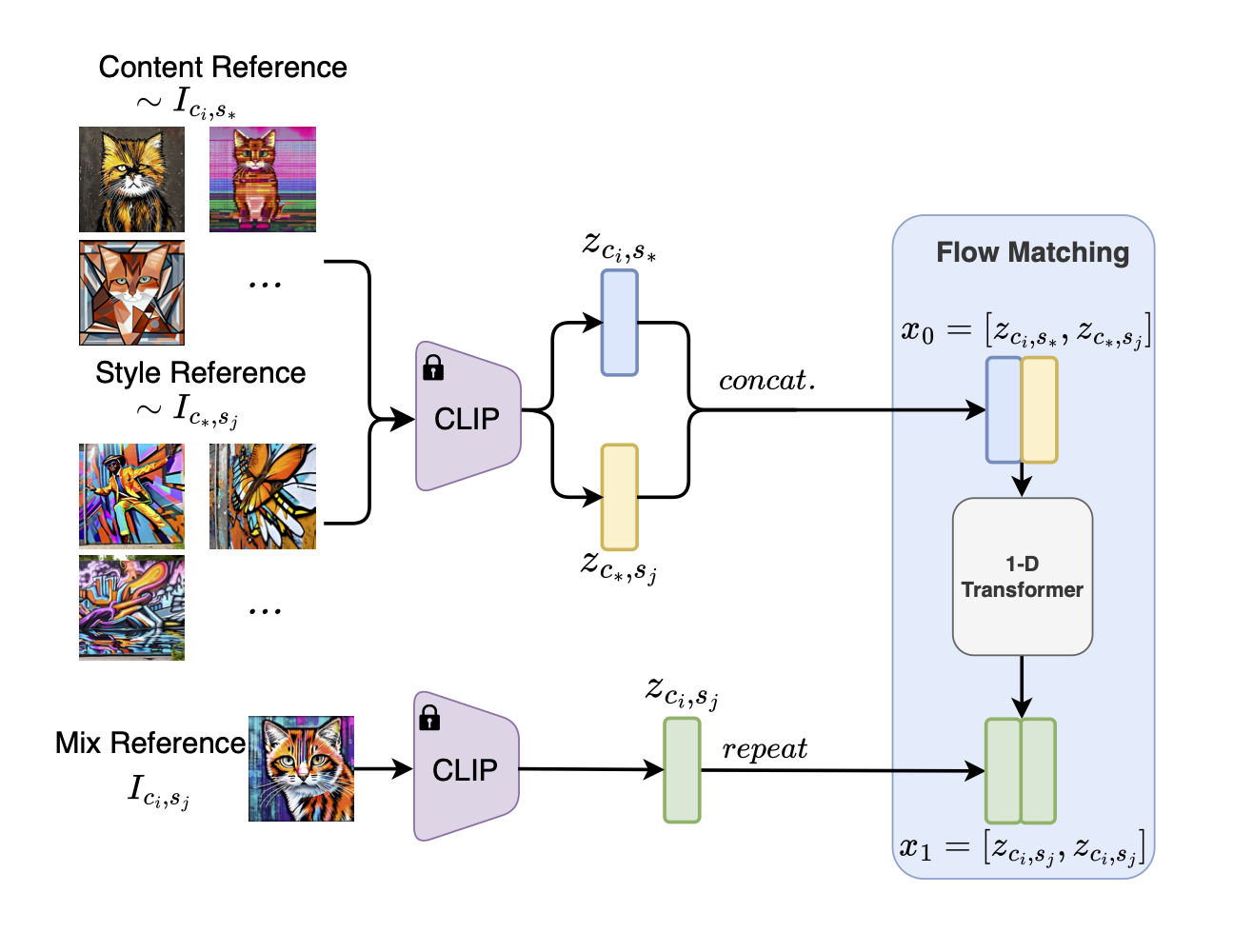}
    \caption{Our training pipeline. $*$ denoted \textit{arbitrary} instance.}
    \label{fig:training}
\end{figure}

\noindent Explicitly defining style (\( s \)) and content (\( c \)) is inherently challenging due to their semantic entanglement and ambiguity. Rather than imposing rigid separation criteria, we propose to learn disentanglement implicitly by modeling the transfer between two data distributions: the disentangled distribution \( p_0(x) \), representing ``pure'' style and content \(x_0 = (c, s)\), and the merged distribution \( p_1(x) \), representing stylized outputs \(x_1=c \oplus s\), where style and content are mixed. Our goal is to learn a bidirectional mapping between these distributions: a blending process (\( p_0 \rightarrow p_1 \)) to merge \( c \) and \( s \), and a disentangling process (\( p_1 \rightarrow p_0 \)) to recover them from \( c \oplus s \). Importantly, training only in one direction suffices if the mapping is invertible.

Flow Matching models~\cite{albergo2023stochastic,ma2024sit, albergo2022building,tong2023improving,lipman2023flow} are perfectly suitable for this task (see \cref{sec:flow_matching}). Unlike diffusion models~\cite{ldm,fuest2024diffusion}, which require Gaussian noise as one of the end distributions, FMs directly learn deterministic paths between \( p_0 \) and \( p_1 \), as long as one can sample from them. 
By training the proposed \textit{SCFlow} solely to blend, \ie, \( p_0 \rightarrow p_1 \), the invertibility allows disentanglement to emerge without explicit supervision given as a separation task.

In practice, however, aligned representations of ``pure'' style/content pairs (\( c, s \)) and their mixed counterparts \( c \oplus s \) are not readily available in existing datasets.  Therefore, we curate a dataset of asymmetric triplets (\( c_is_*, c_*s_j, c_is_j \)) with full combinatorial coverage (see \cref{sec:data_construction}), \textit{where $*$ denotes arbitrary choices.} On one side, the model has access to more information (the \(s_*\) and \(c_*\) in \( c_is_*, c_*s_j\)) than the other side (\(c_is_j\)). This structured asymmetry forces the model to learn invariant attributes while discarding irrelevant variations (see \cref{sec:training}), \eg, learning content from style-varying samples. As a result, the model learns disentanglement by construction, despite never observing explicitly labeled representations of \(c\) or \(s\) (see \cref{sec:inference}).

\subsection{Flow Matching}
\label{sec:flow_matching}

Flow Matching (FM) provides us the principal framework for learning deterministic paths between the disentangled \( p_0(x) \) and the merged \( p_1(x) \) distributions. The mapping between  \( p_0(x) \) and  \( p_1(x) \) can be defined~\cite{albergo2023stochastic} by a time-dependent forward process with $t \in [0, 1]$ as
\begin{equation}
x_t = \alpha_t x_0 + \sigma_t x_1,
\end{equation}
where \( x_0\)  corresponds to sample pairs of \( c_is_*, c_*s_j\) simulating content and style references, 
and \( x_1 \) denotes the merged form \(c_is_j\).
The forward process (blending process \( p_0 \rightarrow p_1 \)) is characterized by the coefficients $\alpha_t$ and $\sigma_t$, with $\alpha_t$ decreasing and $\sigma_t$ increasing as time $t \in [0,1]$ progresses, interpolating between $p_0(x)$ and $p_1(x)$. Furthermore, the boundary conditions are normally constrained as $\alpha_0 = \sigma_1 = 1$ and $\alpha_1 = \sigma_0 = 0$, so that for $t=0$ we have pure data from $p_0(x)$ and $t=1$ we have pure data from $p_1(x)$. 

The velocity, which governs the Ordinary Differential Equation (ODE) dynamics $\frac{dx}{dt}=v(x, t)$, is defined as
\begin{equation}
v(x, t)=\mathbb{E}[\dot{x}_t | x_t = x],
\end{equation}
which generates the marginal probability distribution $p_t(x)$ of $x_t$ at time $t$~\citep{albergo2023stochastic,ma2024sit,song2021scorebased_sde}.

During inference, we solve the probability flow ODE along $t$: with \text{ODESolve} denoting the bidirectional mapping between the disentangled data $x_0$ and the merged data $x_1$ :
\begin{equation}
\label{eq:ode_solve}
    \text{ODESolve}(x_t; v)_{[0, 1]} = x_0 + \int_0^1v(x_t,t)dt,
\end{equation}
to obtained \( x_1 = c \oplus s \) from \( x_0 = (c, s) \). Notably, this process can be done reversely to achieve the \textit{disentangling} (\( p_1 \rightarrow p_0 \)), with the same $v(x_t,t)$ by only changing the direction of the integral operator as $\text{ODESolve}(x_t; v)_{[1, 0]}$.

\citet{ma2024sit} showed that we can train a neural network $v_\theta(x,t)$ to approximate the velocity $v(\cdot,\cdot)$ using the following training objective:
\begin{align}
\label{eq:velocity-eq-obj}
\mathcal{L}(\theta)
&= \int_0^T \E[\Vert v_\theta(x_t, t) - \dot\alpha_t x_0 - \dot\sigma_t x_1\Vert^2] \rd t,
\end{align}
with $\dot{\alpha}_t$ and $\dot{\sigma}_t$ representing the time derivative of $\alpha_t$ and $\sigma_t$, respectively. For simplicity and due to its relatively straight trajectories, we adopt the \textit{Linear schedule} from \cite{ma2024sit,rectifiedflow_iclr23} with $\alpha_t = 1 - t$ and $\sigma_t = t$. This definition inherently leads to optimal transport~\cite{lipman2022flow}. 
However, this path might be suboptimal when the end distributions $p_0(x)$ and $p_1(x)$ are correlated while sample pairs are drawn randomly from them simultaneously.

\begin{figure}[htb]
    \centering
    \includegraphics[width=1\linewidth]{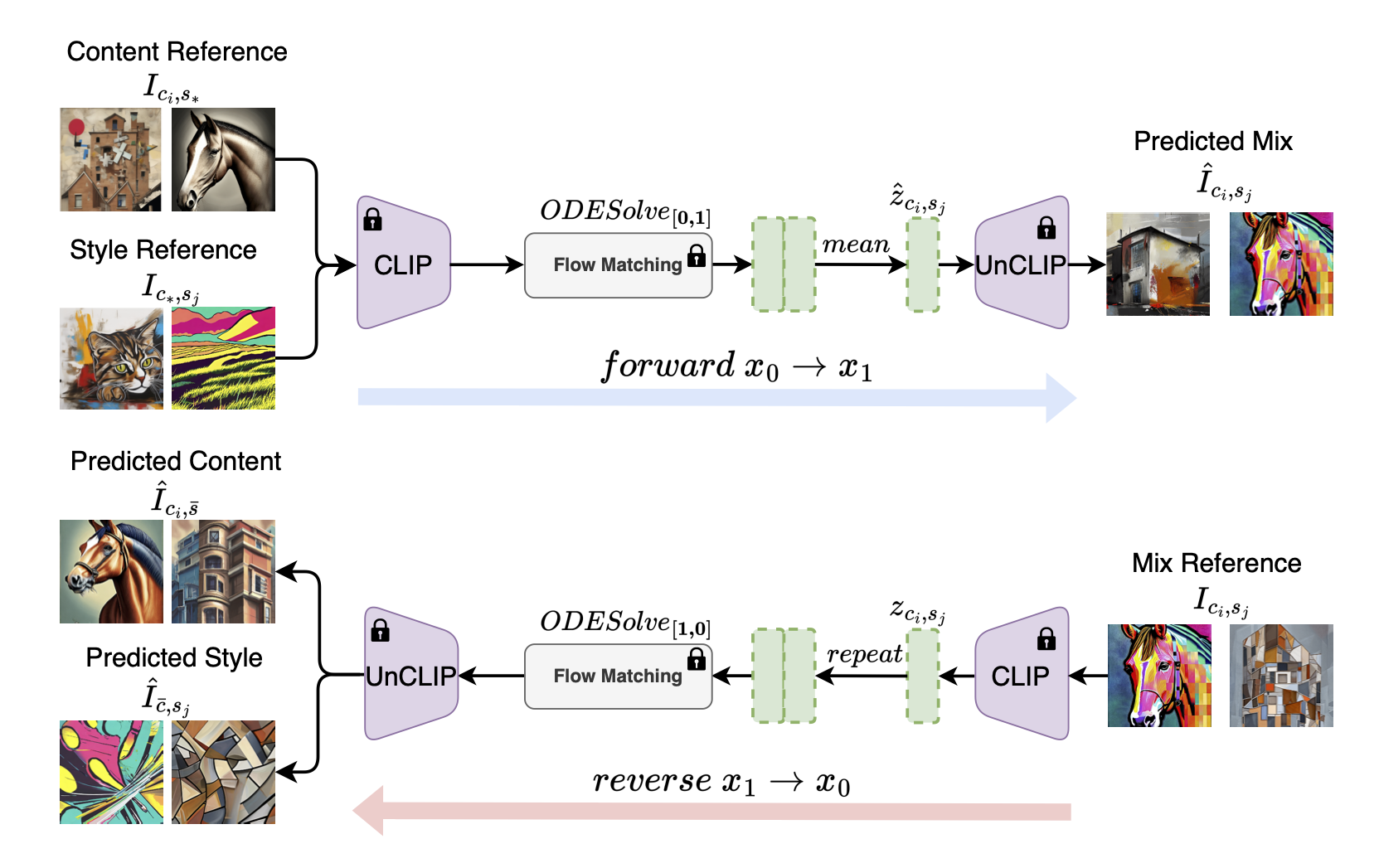}
    \caption{Bidirectional Inference denoted by $[0,1]$ and $[1,0]$.}
    \label{fig:inference}
\end{figure}

\subsection{Matching Style and Content with Dependency}

The naive approach would be to train the $v_\theta$ while \textit{randomly} sampling from $p_0(x)$ and $p_1(x)$. However, this leads to a \textit{moving target problem} as the sampled style and content from $p_0(x)$ do not necessarily introduce the sampled mixture in $p_1(x)$. This problem becomes even more pronounced as the number of distinct styles and contents grows.

To address this, \textit{the key is to sample from these two distributions dependently}, \ie, ensuring the style and content sampled from $p_0(x)$ match the merged counterpart in $p_1(x)$. This dependency allows the model to learn the mapping between the disentangled and mixed distributions effectively. As demonstrated in previous works~\cite{tong2023improving,gui2024depthfm,2023boosting}, such a dependency also enables faster and more stable training. Implementing this dependency necessitates a dataset containing aligned triplets of content $c$, style $s$, and their mixture $c \oplus s$.

\paragraph{Construction of Our Dataset.}
\label{sec:data_construction}

Existing datasets that focus on style, such as WikiArt~\cite{saleh2015large}, BAM~\cite{wilber2017bam}, BAM-FG~\cite{ruta2021aladin}, and LAION-Styles~\cite{somepalli2024CSD}, lack triplets of content $c$, style $s$, and their mixture $c \oplus s$. They primarily consist of individual images labeled with a specific style, without systematic combinations of content and style. Similarly, large-scale datasets capturing diverse content, such as CC3M~\cite{sharma2018conceptual}, LAION~\cite{schuhmann2021laion,laion-5b}, and Unsplash~\cite{unsplash}, do not provide stylized variations of the same content.

To address these limitations, we construct a dataset with full combinatorial coverage of $51$ styles $\times$ $10{,}000$ content instances, resulting in $510{,}000$ samples. Every content instance is paired with every style, enabling systematic analysis of style–content interactions. The original content images are scraped from Pexels, and stylized variants are generated using ControlNet~\cite{zhang2023controlNet}. For details of the construction pipeline, we refer to \cref{sec:app:content_collection}. We split the dataset based on content, with $70\%$ for training and $30\%$ for testing.

Different from existing datasets, we ensure that each style contains images for all possible contents, and each content has all possible styles. This enables us to sample \emph{asymmetric triplets} of the form ${c_{i}s_{*}}$ (content \(i\) with arbitrary style),  ${c_{*}s_{j}}$ (style \(j\) with arbitrary content), and ${c_{i}s_{j}}$ (both style and content fixed). This structured asymmetry encourages the model to disentangle content and style attributes by design.

\subsection{Implicitly Learning Disentangled Representations by Merging}
\label{sec:training}
While the pixel space contains all the necessary information for style and content, its dense spatial details can bias the model toward low-level patterns, easily leading to entanglement and overfitting to irrelevant features. To avoid such spatial biases and emphasize abstract semantics, we design our method to operate in a compact latent space that preserves essential high-level information while discarding redundant low-level variation.

For a semantically rich and compact representation, we use CLIP~\cite{radford2021learning} as our image encoder $\mathcal{E}$, providing a shared image-text embedding space that facilitates evaluation and visualization. The well-explored nature of CLIP allows for straightforward assessment via CLIPScore~\cite{hessel2021clipscore} and easy visualization using unCLIP~\cite{ramesh2022hierarchical,ldm}.

We denote our latent as $z_{c_i,s_j}=\mathcal{E}(I_{c_i,s_j})$, given an image $I_{c_i,s_j}$ with content \(i\) and style \(j\). However, $z_{c_i,s_j}$ contains both style and content information. This violates the above assumptions, where we had a clean data distribution for content and style. To tackle this issue, we define our two terminal distributions as:
\begin{align}
    x_0 = [z_{c_i,s_*}, z_{c_*,s_j}]\sim p_0(x),\\
    x_1 = [z_{c_i,s_j}, z_{c_i,s_j}]\sim p_1(x),
\end{align}
where $x_0$ is a concatenation of two embeddings differing in both style and content, and $x_1$ is a repeated version of a single embedding that contains only half the styles and contents in $x_0$. With such a data-dependent and asymmetrical construction of the input, the model has to perform two tasks to accomplish the merging process: \textbf{1)}~Removing irrelevant information contained in $s_*$ and $c_*$ respectively, as they were not reflected by $x_1$; and \textbf{2)} extracting the useful style $s_j$ and content $c_i$ from the entangled $z_{c_i,s_*}, z_{c_*,s_j}$.
This process is illustrated in \cref{fig:training}. With the defined terminal distribution, we can optimize our model using the loss in \cref{eq:velocity-eq-obj}. Such a construction of the data triplet, \ie, content reference $z_{c_i,s_*}$, style reference $z_{c_*,s_j}$ and the mix reference $z_{c_i,s_j}$, circumvents the problem of explicitly defining \textit{what style and content are within the given representation} and allows the model to discover this implicitly.

\subsection{Inference}
\label{sec:inference}
The trained model only requires samples from one of the distributions and does not rely on additional conditions ~\cite{2023boosting,gui2024depthfm}, so it can perform inference in both directions.
Firstly, we can merge the given style and content reference and remove their irrelevant part as shown in the upper part of \cref{fig:inference}. We define this as the $forward$ process with the ODESolve operator in \cref{eq:ode_solve}:
\begin{equation}
\label{eq:forward}
z_{c_i, s_j} = \text{mean}(\text{ODESolve}([z_{c_i,s_*}, z_{c_*, s_j}])_{[0,1]}),
\end{equation}
where $s_*$ and $c_*$ denote any arbitrary styles and contents. The style in the content reference does not matter, and vice versa (see examples in \cref{{fig:mix_vis}}). 

More interestingly, with the model only trained for the $forward$ process, we can perform the inference from the other direction, due to the invertibility offered by flow models~\cite{tong2023improving,taohu2023lfm,ma2024sit}. We define this direction as $reverse$, illustrated on the lower part of \cref{fig:inference}:
\begin{equation}
\label{eq:reverse}
[z_{c_i,\bar{s}}, z_{\bar{c}, s_j}]= \text{ODESolve}(\text{repeat}[z_{c_i, s_j}])_{[1,0]},
\end{equation}
where $\bar{s}$ and $\bar{c}$ denote the mean values of styles and contents under the condition giving $c_i, s_j$ across the dataset and $\text{repeat}$ duplicates the input twice to match the dimension, \ie, $\text{repeat}[z_{c_i, s_j}] = [z_{c_i, s_j}, z_{c_i, s_j}]$. If not specified elsewhere, we use $1$ as the number of function evaluations (NFE) by default.

\label{sec:exp}
\begin{figure*}[ht]
    \centering
    \includegraphics[width=1\linewidth, height=1\linewidth, keepaspectratio]{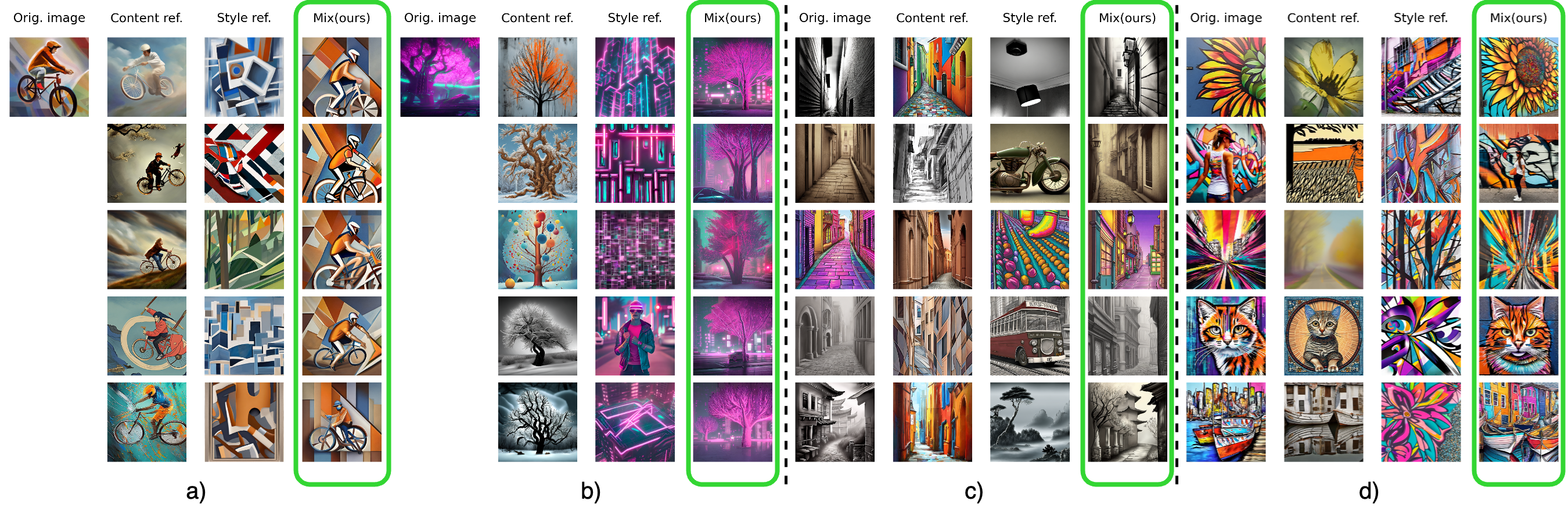}
    \caption{Visual results of forward inference $x_0 \to x_1$. The first column shows the original image with the targeted content and style. The second and third columns show the respective content and style references. \textit{a,b)} generated results from content $z_{c_i,s_*}$ and style  $z_{c_*,s_i}$ references. \textit{c)} We keep the content fixed but with varying styles. \textit{d)} The style is fixed, while we change the contents.}
    \label{fig:mix_vis}
    \vspace{-8pt}
\end{figure*}

For two reasons, we empirically found that $\bar{s}$ and $\bar{c}$ resemble the dataset mean for style and content. Firstly, they are unaffected by the original content and style respectively, \ie, the original content does not affect $\bar{s}$ and similarly between original style and $\bar{c}$. Secondly, across different combinations of style and content, $z_{c_*,\bar{s}}$ is shown to be almost equidistant to the centroids of any known styles. The same holds analogously for $z_{\bar{c},s_*}$, see details in \cref{sec:app:mean_content_style}.

\section{Experiments}
\subsection{Qualitative Analysis}
\label{sec:experiemtns_our_data}

\paragraph{Forward Inference.}
The mixed latent representations are obtained via forward inference, $x_0 \to x_1$, and visualized using unCLIP~\cite{ramesh2022hierarchical,ldm} to produce the predicted combination $\hat{I}_{c_i, s_j}$. We evaluate three scenarios that highlight the model’s ability to disentangle and combine content and style (\cref{fig:mix_vis}~\textit{a--d}). In the first scenario, we combine \( z_{c_i, s_*} \) (content \(i\), arbitrary style) with \( z_{c_*, s_j} \) (arbitrary content, style \(j\)). Despite the arbitrary components, the output consistently reflects the intended combination \( z_{c_i, s_j} \), demonstrating that the model isolates and fuses the specified content and style while ignoring irrelevant variations (\cref{fig:mix_vis}~\textit{a, b}).

We explicitly vary one factor in the second and third scenarios while fixing the other. In \cref{fig:mix_vis}~\textit{c}, we fix the content reference \( z_{c_i, s_*} \) and iterate the style reference \( z_{c_*, s_j} \) over \(j\), producing outputs that consistently depict content \(i\) rendered in each style \(j\). Conversely, in \cref{fig:mix_vis}~\textit{d}, we fix the style reference \( z_{c_*, s_j} \) and iterate the content reference \( z_{c_i, s_*} \) over \(i\), generating outputs that show different contents \(i\) rendered in the fixed style \(j\). Across all cases, the model disentangles content and style semantics solely from references in the embedding space \textit{without any other guidance, \eg text,} and combines them effectively when decoding with unCLIP.

\vspace{-10pt}

\paragraph{Reverse Inference.}
During reverse inference, $x_1 \to x_0$, the model produces disentangled latent representations of content and style. As shown in \cref{fig:vis_backward}, the predicted content exhibits no stylistic influence from the original mixed reference, and the predicted style is abstracted from any specific content, capturing only high-level stylistic patterns. These results demonstrate the model’s ability to isolate and extract pure content and style semantics effectively. Additionally, we visualize the aggregated latent proxies in \cref{sec:app:avg_content} and discuss the mean content $\bar{c}$ and style $\bar{s}$ carried by the disentangled representation in \cref{sec:app:mean_content_style}.

\begin{figure}[h]
    \centering
    \includegraphics[width=0.6\linewidth]{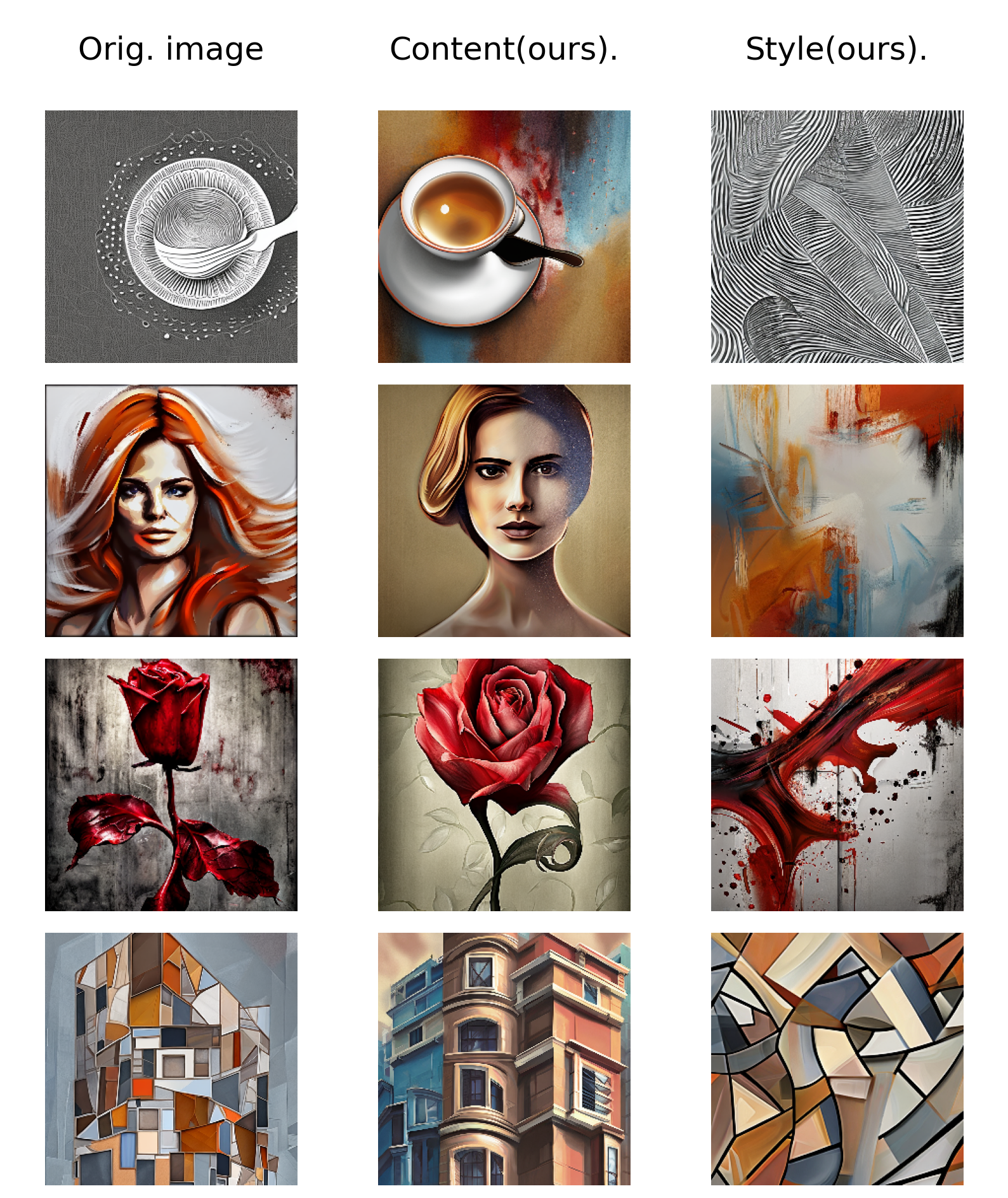}
    \caption{Visual results of reverse inference $x_1 \to x_0$. The first column shows the mix references $z_{c_i, s_j}$; The second and third columns show the predicted content and style.}
    \label{fig:vis_backward}
    \vspace{-10pt}
\end{figure}

\subsection{Evaluation of Latent Representations}

\begin{figure}[h]
    \centering
    \includegraphics[width=\linewidth]{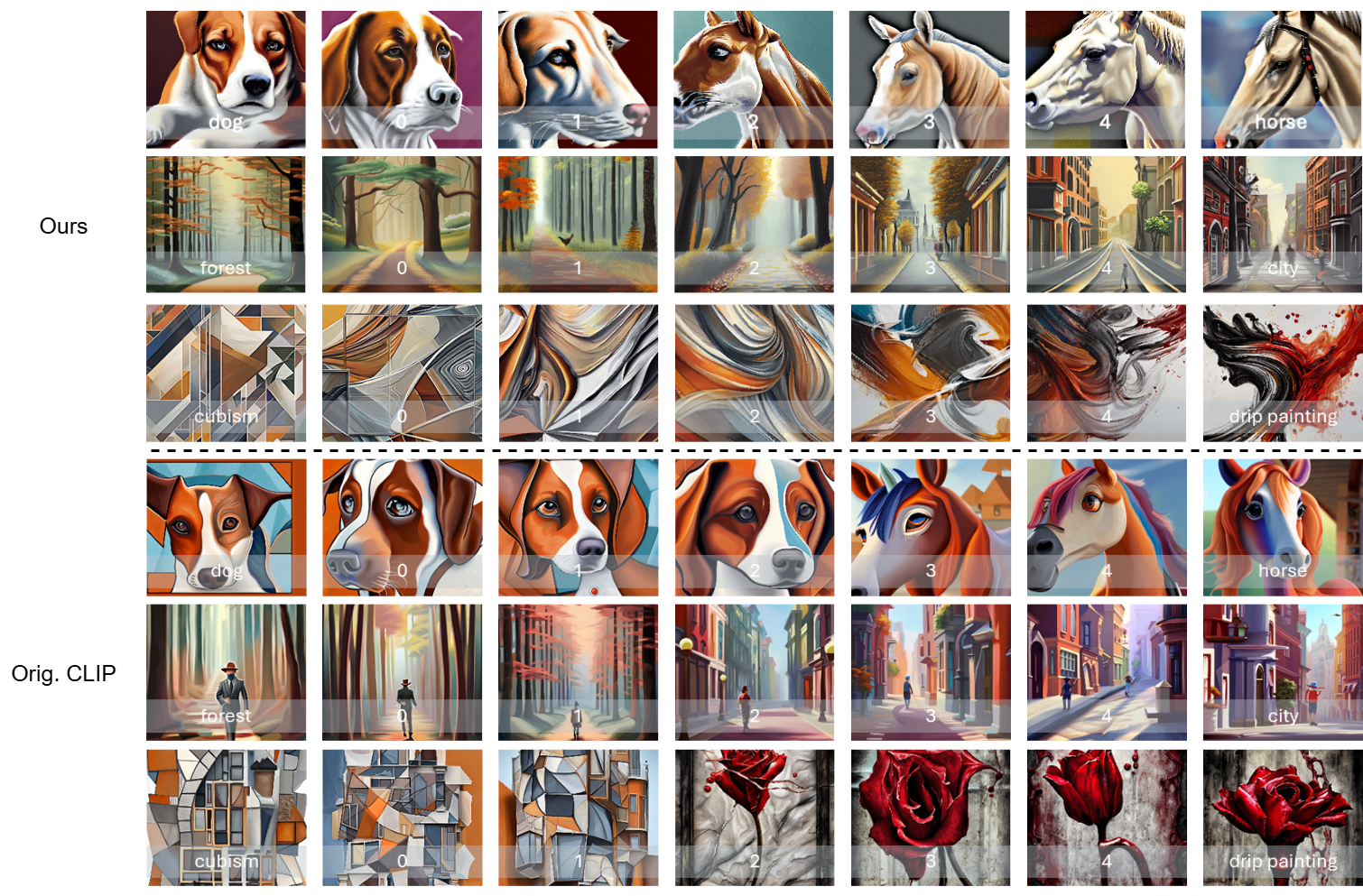}
    \caption{Visualization of the linear interpolation. The interpolation was done between pairs of contents and styles embedding.
    }
    \vspace{-10pt}
    \label{fig:i_to_j}
\end{figure}

\paragraph{Disentanglement of Style and Content Representations.}

We assess the quality of the learned content and style embeddings by visualization and quantitative metrics. We first visualize the embedding spaces using t-SNE~\cite{van2008tsne} in \cref{fig:tsne_style} and \cref{fig:tsne_content}, where 25 randomly selected content and style classes are shown. Compared to the original CLIP~\cite{radford2021learning} space, our embeddings form more compact and well-separated clusters, with semantically similar classes positioned closer together. This structure is further illustrated in the right-hand sections of \cref{fig:tsne_style} and \cref{fig:tsne_content}, where decoded images from the mean embeddings of selected classes are overlaid on the plots. Classes with larger semantic differences are placed farther apart without explicit constraints, reflecting better organization of the latent space.

To quantify these observations, we apply K-means~\cite{macqueen1967some} clustering and compute the normalized mutual information (NMI)~\cite{manning2009NMI} between the predicted clusters and ground-truth labels. As shown in \cref{tab:nmi_scores}, our model achieves the highest NMI scores for both content and style, significantly outperforming CLIP, DEADiff~\cite{Qi_2024}, and CSD~\cite{somepalli2024CSD}. While CSD achieves strong style NMI, as it was explicitly trained for style representation, it fails on content disentanglement.
We also report the Fisher Discriminant Ratio (FDR)~\cite{fisher1936use}, which measures inter-class versus intra-class variance, with higher values indicating better separability. Our embeddings achieve the highest FDR for content and style, with style FDR an order of magnitude higher than CLIP and DEADiff, and five times higher than CSD. We refer to \cref{sec:app:eval_details} for obtaining their embeddings.

Finally, we jointly visualize randomly selected content and style classes in \cref{fig:c_n_s} to compare the disentanglement. Our embeddings exhibit clear separation between content and style, while CLIP embeddings show significant overlap, likely due to their entangled representation of both factors. This demonstrates that our method produces more structured and disentangled latent spaces than prior approaches. Additional experiments in \cref{sec:app:contrastive} show that our performance gains persist even when compared against contrastive learning baselines~\cite{chopra2005learning,oord2018representation} trained on the same dataset, confirming the advantage of our method beyond data alone.

\vspace{-10pt}

\begin{figure}[ht] 
    \centering
    \includegraphics[width=1\linewidth]{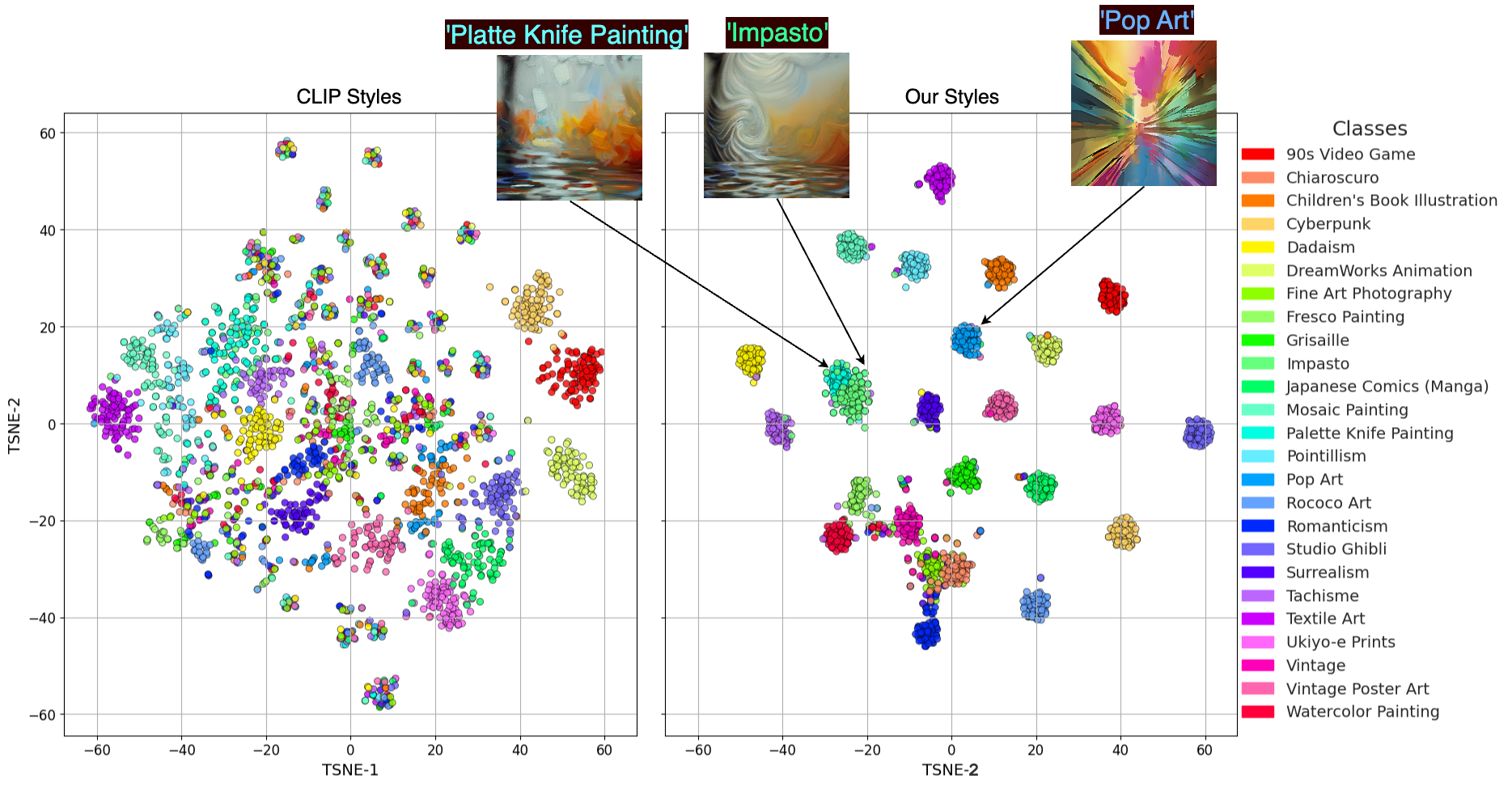}
    \caption{t-SNE comparison of style. The same set of styles for both methods was randomly selected from the test split.} 
    \label{fig:tsne_style}
    \vspace{-10pt}
\end{figure}

\begin{figure}[h] 
    \centering
    \includegraphics[width=1\linewidth]{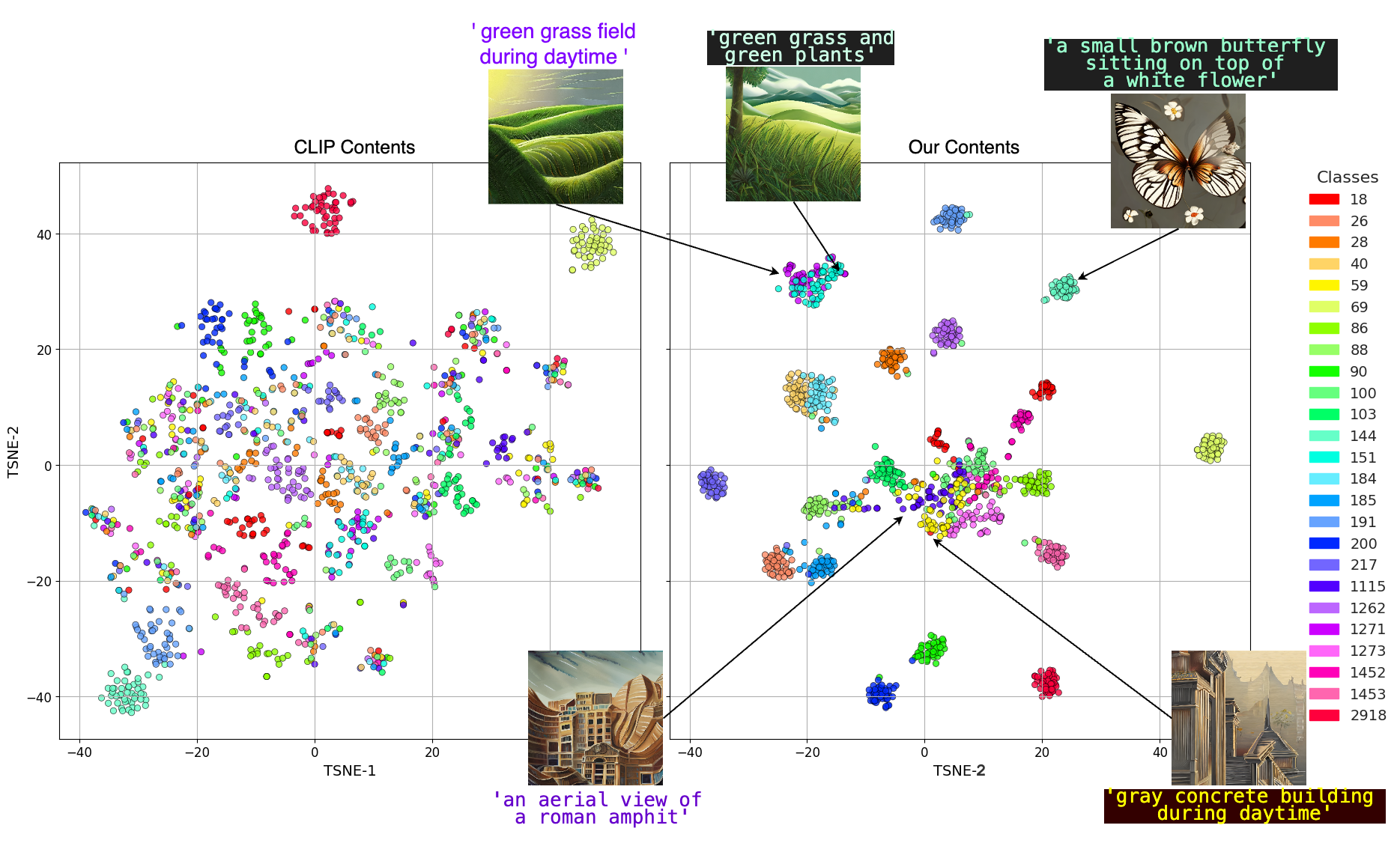}
    \caption{t-SNE comparison of content. The same set of contents for both methods was randomly selected from the test split.}
    \vspace{-10pt}
    \label{fig:tsne_content}
\end{figure}

\begin{figure*}[ht]
    \centering
    \begin{minipage}[t]{0.23\textwidth}
        \centering
        \captionsetup{labelformat=empty}
        \scriptsize
        \vspace{-2.5cm} 
        \begin{tabular}{lccc}
            \toprule
            \textbf{Similarity$\uparrow$} & \multicolumn{3}{c}{\textbf{Pairs}} \\
            & D, H & F, C  & C, DP  \\
            \midrule
            CLIP      & 0.13 & 0.29 & 0.25\\
            \textit{Ours} & \textbf{0.38} & \textbf{0.54} & \textbf{0.49}\\
            \bottomrule
            \end{tabular}
    \end{minipage}
    \hfill
    \begin{minipage}[b]{0.24\textwidth}
        \centering
        \includegraphics[width=\textwidth]{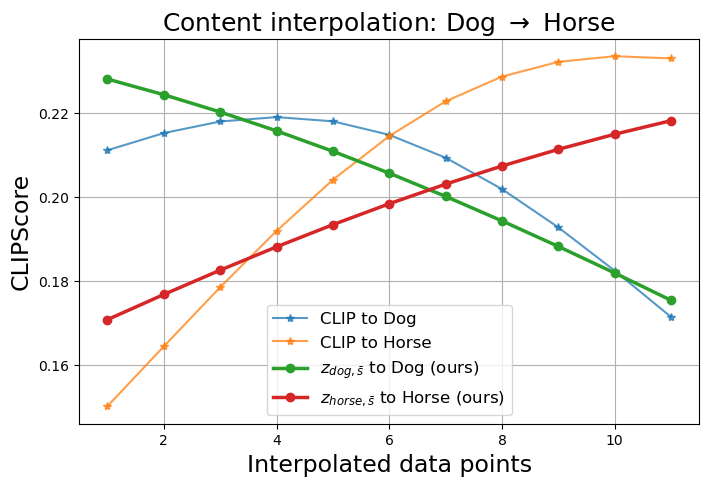}
    \end{minipage}
    \hfill
    \begin{minipage}[b]{0.24\textwidth}
        \centering
        \includegraphics[width=\textwidth]{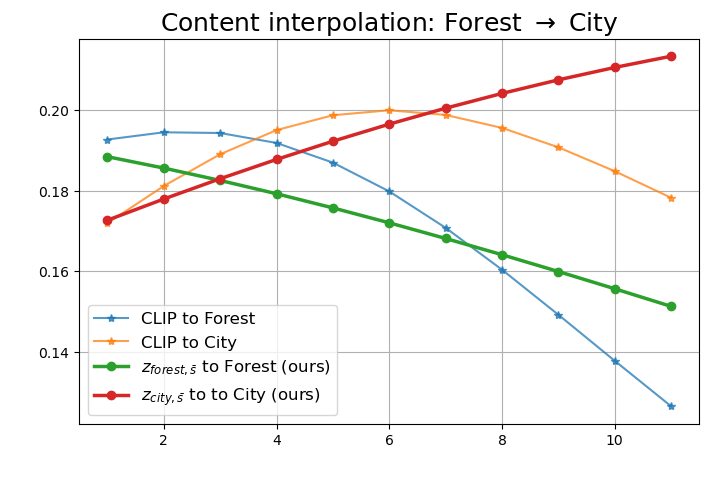}
    \end{minipage}
    \hfill
    \begin{minipage}[b]{0.24\textwidth}
        \centering
        \includegraphics[width=\textwidth]{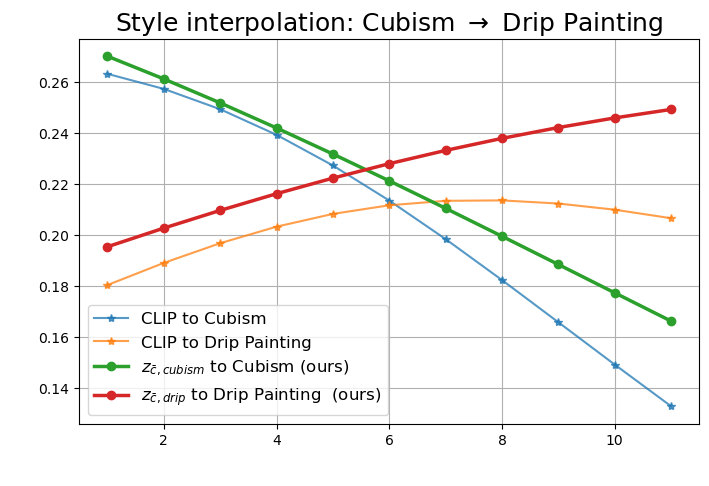}
    \end{minipage}
    \caption{CLIP score of the intermediate interpolated data. \textit{Left).} Cosine similarity between trajectory for text and image embeddings for each pair of concepts (same as \cref{fig:i_to_j}) to assess their alignment with text. \textit{Right).} {\color{ForestGreen}Green} and {\color{Red}red} lines represent our method. 
    }
    \vspace{-10pt}
    \label{fig:clipscore}
\end{figure*}

\begin{table}[h]
    \centering
    \resizebox{\linewidth}{!}{
    \begin{tabular}{c||c|c|c|c|c|c|c|c}
        \hline
        \multirow{2}{*}{\textbf{Clusters}} & \multicolumn{4}{c}{\textbf{NMI Score}$\uparrow$} & \multicolumn{4}{c}{\textbf{FDR$\uparrow$}} \\ 
        & Ours &  CLIP & DEADiff\cite{Qi_2024} & CSD\cite{somepalli2024CSD} & Ours &  CLIP & DEADiff & CSD\\ 
        \midrule
        \rowcolor{gray!8}10 styles  & \textbf{0.8143}  & 0.4888 & 0.3894 & 0.6936 & \textbf{2.5976} & 0.2353 & 0.2601 & 0.5150 \\  
        25 styles  & \textbf{0.9202}  & 0.5838 & 0.5083 & 0.8229 & \textbf{3.7271} & 0.3066 & 0.3361 & 0.6658 \\  
        \rowcolor{gray!8}51 styles  & \textbf{0.8696}  & 0.4016 & 0.4136 & 0.7241 & \textbf{3.5184} & 0.2961 & 0.3379 & 0.6328 \\  
        \midrule
        25 contents  & \textbf{0.8374}  & 0.3676 & 0.6459 & 0.1888 & \textbf{1.7998} & 0.3340 & 0.4386 & 0.2686 \\  
        \rowcolor{gray!8}200 contents  & \textbf{0.8356}  & 0.5368 & 0.5056 & 0.3345 & \textbf{2.1693} & 0.4307 & 0.5574 & 0.3083 \\  
        \bottomrule
    \end{tabular}
    }
    \caption{Quantitative comparison. We calculated NMI scores~\cite{manning2009NMI} and FDR~\cite{fisher1936use} for different cluster sizes with K-means~\cite{macqueen1967some}.}
    \label{tab:nmi_scores}
\end{table}

\begin{figure}[]
    \centering
    \includegraphics[width=1\linewidth]{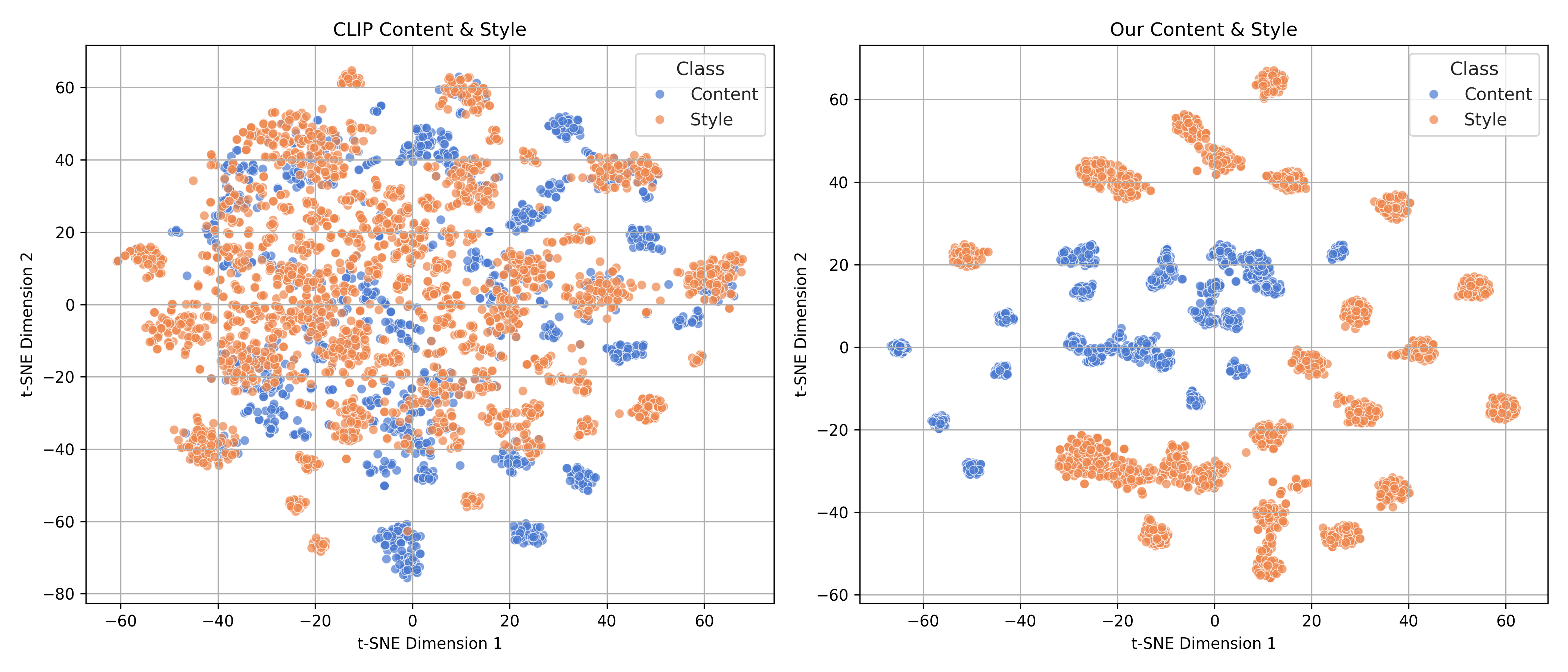}
    \caption{
    t-SNE of content and style embeddings using 25 randomly selected classes from the test set.
    }
    \label{fig:c_n_s}
\end{figure}

\paragraph{Smooth Interpolation in Pure Latent Spaces.}
We examine the smoothness of transitions in our disentangled content and style embeddings by linearly interpolating between two latent vectors, $\lambda z_{c_i} + (1-\lambda)z_{c_j}$ with $\lambda \in [0,1]$. As visualized with unCLIP~\cite{ramesh2022hierarchical} in \cref{fig:i_to_j}, our embeddings produce continuous and semantically meaningful interpolations, whereas the original CLIP~\cite{radford2021learning} yields more abrupt transitions. In \cref{fig:i_to_j}, we interpolate between pairs of content and style latents, \eg, dog–horse (D, H)  or forest–city (F, C) for content, cubism–drip painting (C, DP) for style, using both our method and CLIP. Our content interpolations evolve gradually in the top two rows while maintaining minimal stylistic artifacts; CLIP shows discontinuities. Similarly, in the bottom row, our style interpolation introduces stylistic features progressively, while CLIP exhibits noticeable jumps (\eg, between columns 3 and 4), highlighting better coherence and disentanglement in our latent space.

We further validate these observations quantitatively in \cref{fig:clipscore}. On the left, we compute cosine similarity between the difference vectors in the image embedding space, \ie, $z_{c_i} - z_{c_j}$, and corresponding text embeddings. Our content and style representations (highlighted in bold) align more strongly and consistently with text than CLIP. On the right, we plot CLIP scores~\cite{hessel2021clipscore} along interpolation steps: for our embeddings, the scores decrease smoothly for concept $i$ (green) and increase for concept $j$ (red), reflecting a consistent semantic transition. In contrast, CLIP’s scores (blue and orange) fluctuate irregularly, indicating less interpretable interpolations. These results demonstrate that our disentangled latent spaces enable more natural, coherent, and interpretable transitions than those of CLIP. 

\subsection{Generalization to Unseen Contents and Styles}
\label{sec:imagenet_wikiart}

\begin{table}
    \footnotesize
    \centering
    \begin{tabular}{l||ccc}
    \toprule
    \textbf{Method} & \multicolumn{3}{c}{\textbf{Acc (\%) at \# Neighbors}} \\
    & \textit{$k=1$} & \textit{$k=5$} & \textit{$k=10$} \\
    \midrule
    DEADiff~\cite{Qi_2024}      & 62.81  & 65.69  & 67.07  \\
    \rowcolor{gray!8}CLIP~\cite{radford2021learning} & \textbf{67.10}  & \textbf{69.71}  & \textbf{70.44}  \\
    CSD-C~\cite{somepalli2024CSD} & 56.54  & 59.22  & 61.14  \\
    
    \midrule
    
    \rowcolor{gray!8}\textit{Ours} & \underline{66.25} & \underline{68.74} & \underline{69.67} \\
    \bottomrule
    \end{tabular}
    \caption{Evaluation of content representation on ImageNet-1k~\cite{deng2009imagenet} with k-Nearest Neighbor (kNN) classification 
    }
    \label{tab:ImageNet1k}
\end{table}

\begin{table}
    \footnotesize
    \centering
    \begin{tabular}{l||ccc}
    \toprule
    \textbf{Method} & \multicolumn{3}{c}{\textbf{Recall@$k$}(\%)} \\
    & \textit{$k=1$} & \textit{$k=10$} & \textit{$k=100$} \\
    \midrule
    DEADiff~\cite{Qi_2024}  & 61.24  & 84.09  & \underline{95.67} \\
    \rowcolor{gray!8}CLIP~\cite{radford2021learning} & 59.40  & 82.90  & 95.10  \\
    CSD-S~\cite{somepalli2024CSD} & \underline{64.56} & \underline{85.73}  & 95.58  \\
    \midrule
    \rowcolor{gray!8}\textit{Ours} & \textbf{65.34} & \textbf{88.31} & \textbf{97.67} \\
    \bottomrule
    \end{tabular}    \caption{Recall@$k$~\cite{jegou2010product} on WikiArt~\cite{saleh2015large}} 
    \label{tab:WikiARt}
\end{table}

\paragraph{ImageNet Classification using kNN.}
To evaluate the quality of the extracted content features, we perform \textit{zero-shot k-nearest neighbor classification}~\cite{fix1985discriminatory} on ImageNet-1k~\cite{deng2009imagenet} (\cref{tab:ImageNet1k}). We randomly sample $10\%$ of the training data per class (130 samples per class) and use the whole test set for evaluation. While CSD~\cite{somepalli2024CSD} fine-tunes CLIP embeddings for style, it sacrifices content discrimination and performs poorly here. DEADiff~\cite{Qi_2024} performs reasonably well, even using mean embeddings. Our method achieves content classification performance comparable to the original CLIP, which remains slightly higher, but struggles with style representation, as discussed next.

\paragraph{WikiArt Style Retrieval.}
We assess the style embeddings via \textit{Recall@$k$}~\cite{jegou2010product} on WikiArt~\cite{saleh2015large} (\cref{tab:WikiARt}), following the evaluation protocol in~\cite{somepalli2024CSD}. We split the data into $20/80$ query/gallery sets based on artist labels, repeated over five random splits due to lack of predefined splits, and report the mean Recall. Our method outperforms CLIP in style retrieval while maintaining competitive content classification performance, demonstrating superior disentanglement. CSD ranks second in style retrieval, ahead of DEADiff, as it was explicitly optimized for compact style representation. 

\noindent Overall, our approach performs strongly on content and style, generalizing to unseen styles \textit{without explicit disentanglement objectives}. We further tested our method on previously unseen styles and contents, including unseen textual conditions and real-world samples from ImageNet and WikiArt, shown in \cref{sec:app:unseen}, confirming its ability to extract, fuse, and disentangle new content and style.
\section{Conclusion and Future Work}
In this paper, we introduce a novel method for implicitly disentangling style and content within a semantic space. By utilizing flow matching and a large-scale, curated dataset of content–style pairs, our approach effectively separates style and content from mixtures and generalizes well to unseen data. Extensive experiments demonstrate the effectiveness of our \textit{SCFlow} in both merging and disentangling tasks.

Beyond style and content, we believe this framework has the potential to extend to other abstract modalities and broader applications. In particular, the use of flow matching for mapping between two real data distributions from both directions remains underexplored and presents a promising direction for future work.

\section*{Acknowledgement}
This project was supported by Bayer AG, the Federal Ministry for Economic Affairs and Energy within the project ``NXT GEN AI METHODS - Generative Methoden für Perzeption, Prädiktion und Planung'', the project ``GeniusRobot'' (01IS24083) funded by the Federal Ministry of Research, Technology and Space (BMFTR), and the bidt project KLIMA-MEMES.
The authors gratefully acknowledge the Gauss Center for Supercomputing for providing compute through the NIC on JUWELS/JUPITER at JSC and the HPC resources supplied by the NHR @FAU Erlangen.
}
{
    \small
    \bibliographystyle{ieeenat_fullname}
    \bibliography{main}
}

\clearpage
\appendix
\setcounter{page}{1}
\maketitlesupplementary

\renewcommand{\thetable}{S\arabic{table}}
\renewcommand{\thefigure}{S\arabic{figure}}

\section{Dataset Construction}
\label{sec:app:content_collection}

We curate the original content images from Pexels\footnote[1]{https://www.pexels.com/}, following standard web-scraping practices\footnote[2]{https://huggingface.co/datasets/opendiffusionai/pexels-photos-janpf}. Since the Pexels images often have sparse or incomplete captions, we generate improved captions using LLaVA 1.5~\cite{liu2024improved}.

For the style prompts, we select $51$ artistic styles, such as \eg Cyberpunk and Cubism, each accompanied by a brief explanatory description. The selection was guided by a few art experts, and the descriptions were refined with assistance from ChatGPT-4o~\cite{openai2024gpt4o}.

During stylization, we minimize pixel-level constraints by conditioning on scribbles and applying a tailored guidance scale. We also reweight the style component to ensure strong adherence to the specified artistic style. To generate stylized images, we use ControlNet~\cite{zhang2023controlNet} with prompts in the format:
\begin{center}
\texttt{``An image depicting \{content\_caption\}, in the style of \{style\_prompt\}''}
\end{center}

\begin{figure*}[htbp]
    \centering
    \includegraphics[width=\linewidth]{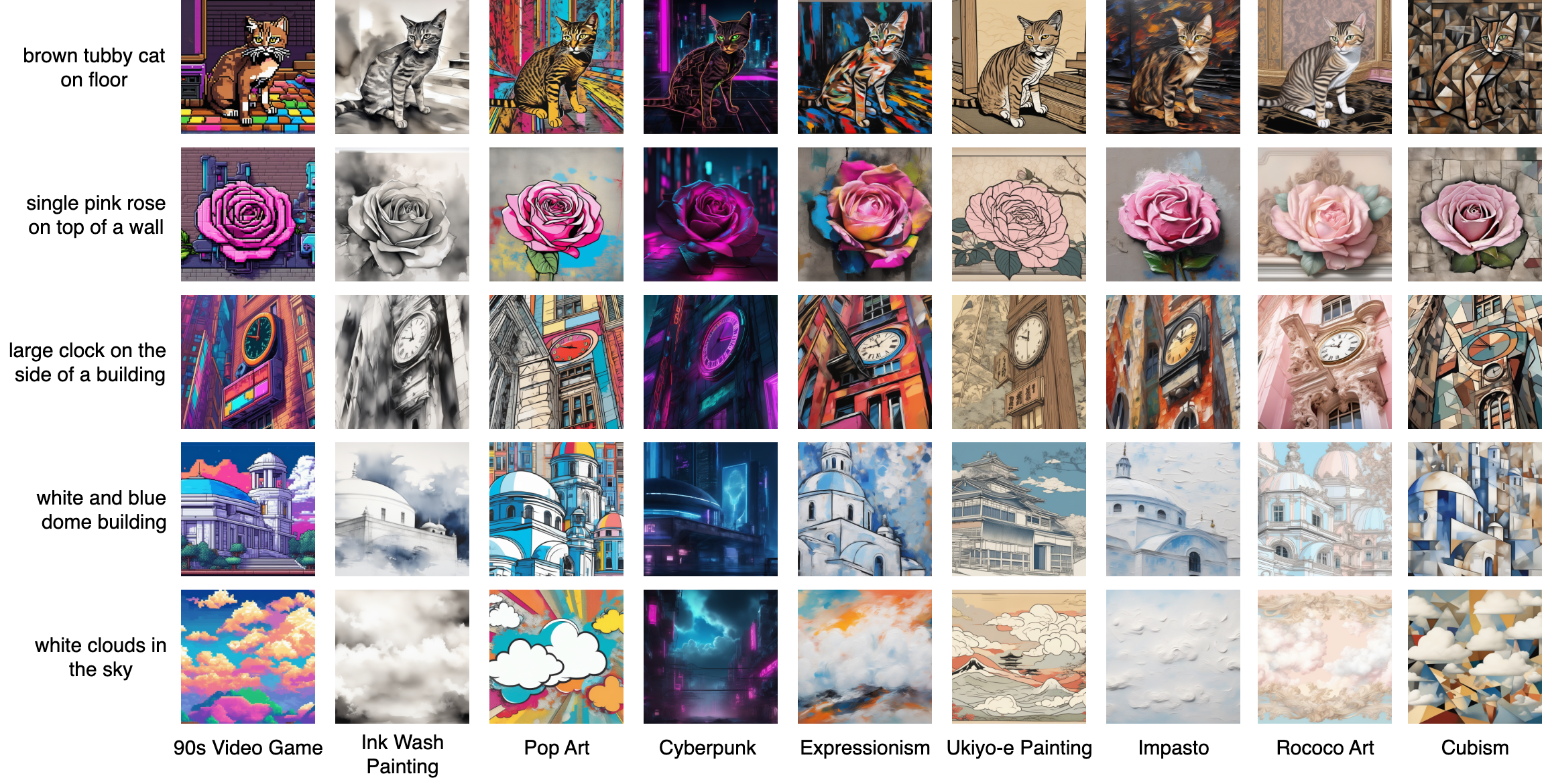}
    \caption{Overview of the curated dataset}
    \label{fig:supp:dataset}
\end{figure*}

We will publish the content captions and the style prompts together with the stylized images. An overview of the curated dataset can be found in \cref{fig:supp:dataset}. Although we use Pexels images as content, the construction pipeline can be easily adapted to other content sources, such as LAION~\cite{laion-5b,schuhmann2021laion} or COYO~\cite{kakaobrain2022coyo-700m}.

\section{Evaluation Details for other models}
\label{sec:app:eval_details}
For CSD~\cite{somepalli2024measuring}, there are two output heads for the style vector and the content vector. Hence, we denote them as CSD-C for content and CSD-S for styles. And we used them accordingly for our evaluation of content and styles. 

For DEADiff~\cite{Qi_2024}, mean query embeddings can be extracted using a pre-trained Q-Former, with visual features corresponding to the prompt ``content'' or ``style''.

\section{Visualization of Content and Style Proxies}
\label{sec:app:avg_content}
\begin{figure}[h]
    \centering
    \includegraphics[width=0.75\linewidth]{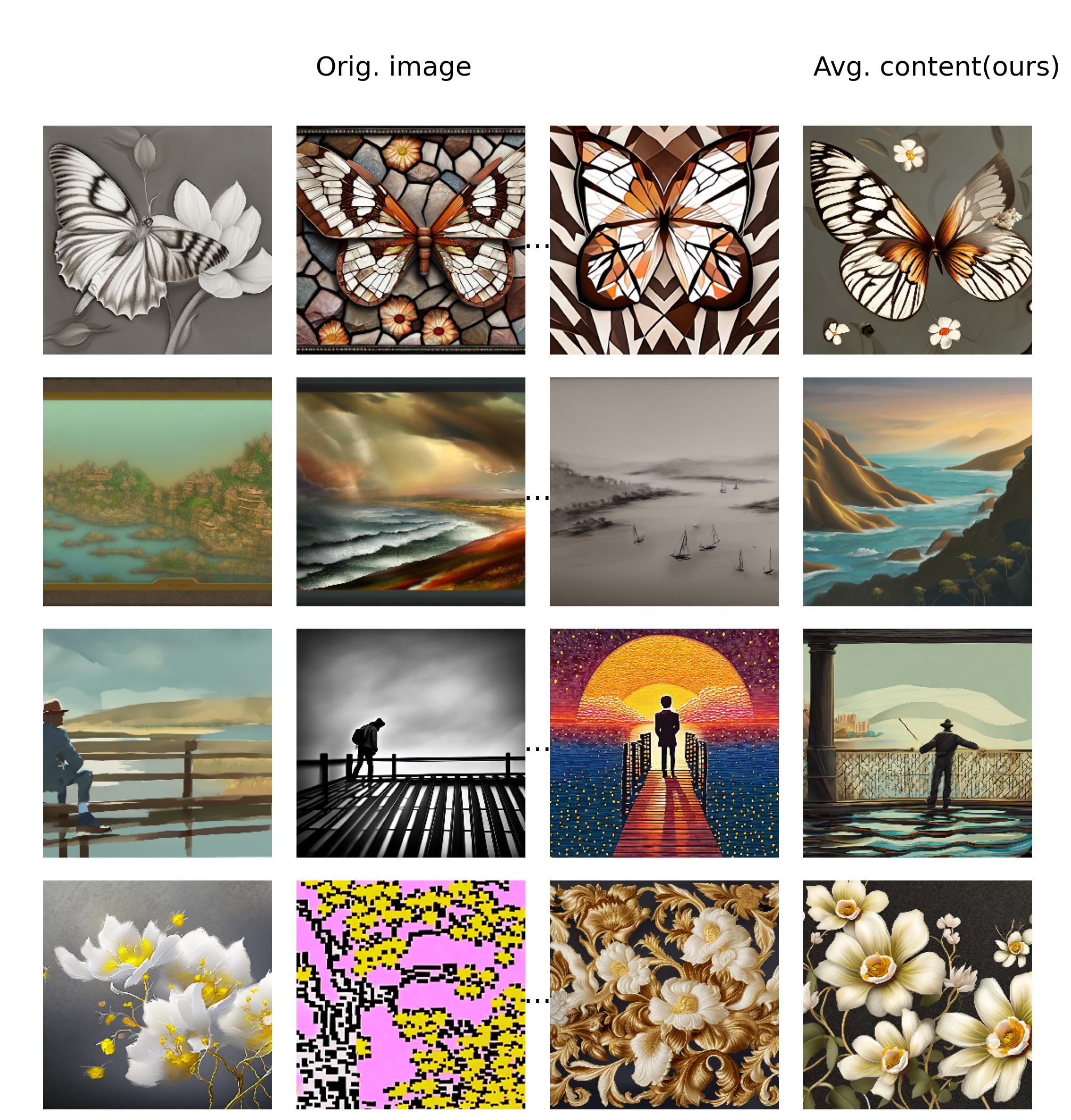}
    \caption{Visualization of proxy contents: The first three columns display a part of the mixed references $I_{c_i, s_j}$, while the last column shows the average content (ours).}
    \label{fig:vis_backward_avg_content}
\end{figure}

\begin{figure}[h]
    \centering
    \includegraphics[width=0.75\linewidth]{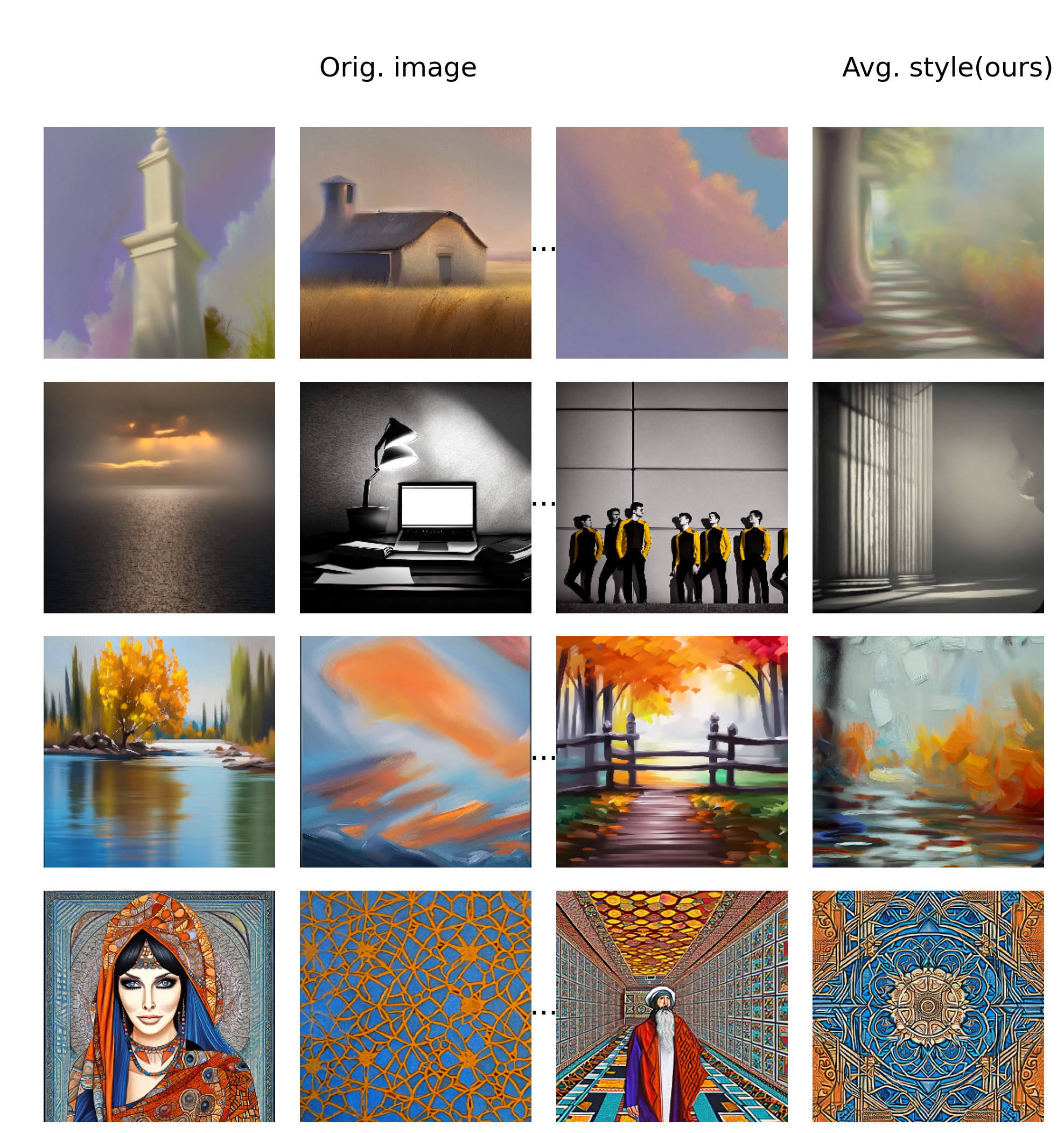}
    \caption{Visualization of proxy styles: The first three columns display a part of the mixed references $I_{c_i, s_j}$, while the last column shows the average style (ours).}
    \label{fig:vis_backward_avg_style}
\end{figure}

We show the aggregated embeddings by averaging them across all predictions conditioned on either style or content, respectively (see \cref{fig:vis_backward_avg_content} and \cref{fig:vis_backward_avg_style}). These aggregated embeddings can be considered as the style or content class proxies in the resulting space.

\begin{figure}[htbp]
    \centering
    \includegraphics[width=.75\linewidth]{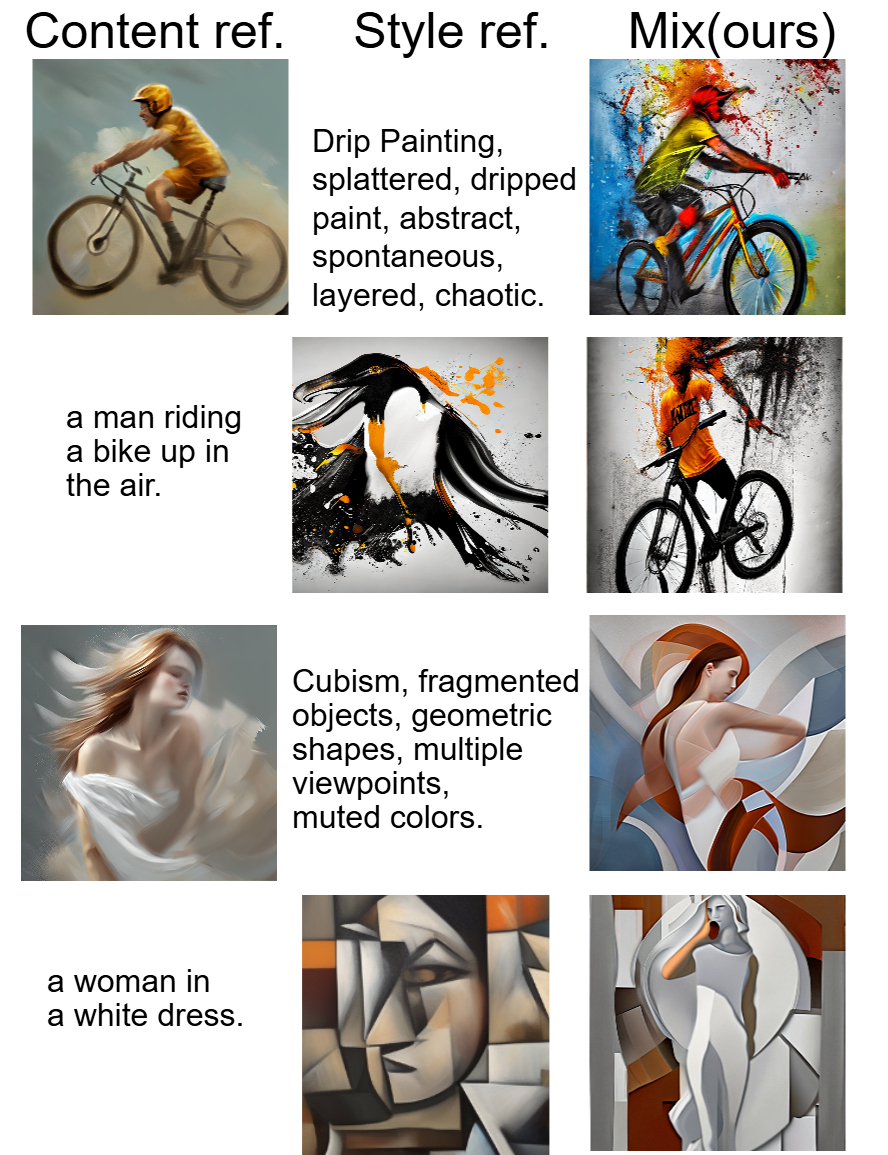}
    \caption{Generalization to textual condition during inference.}
    \label{fig:text_d7P9_3_PH7R_m1}
\end{figure}

\begin{figure}[h]
    \centering
        \centering
        \includegraphics[width=\linewidth]{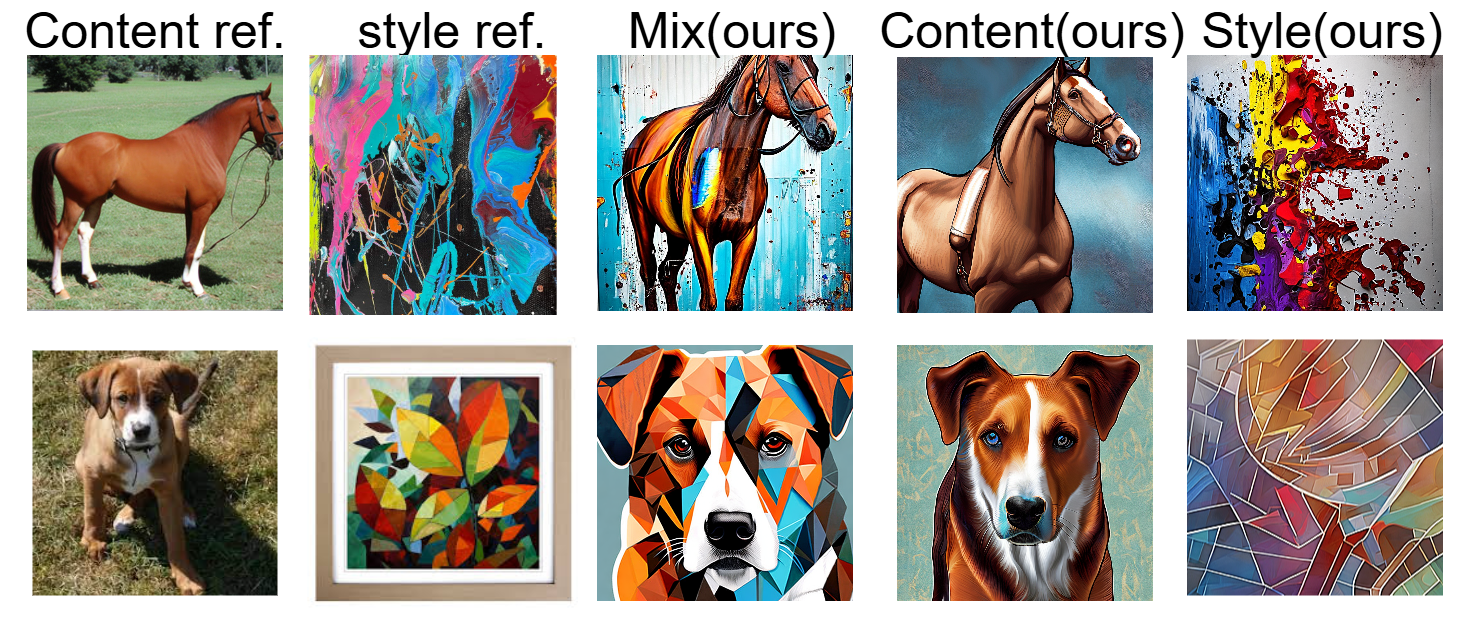}
        \caption{Inference in-the-wild.
        We use ImageNet images as styleless inputs for the forward pass, with style references obtained online. After obtaining the merging results (3rd column), we further use the reverse mapping to map them back to our content and style (4th and 5th columns)
        }
        \label{fig:ImageNet_PH7R_3_d7P9_2}

\end{figure}

\begin{figure}[htbp]
    \centering
    \includegraphics[width=\linewidth]{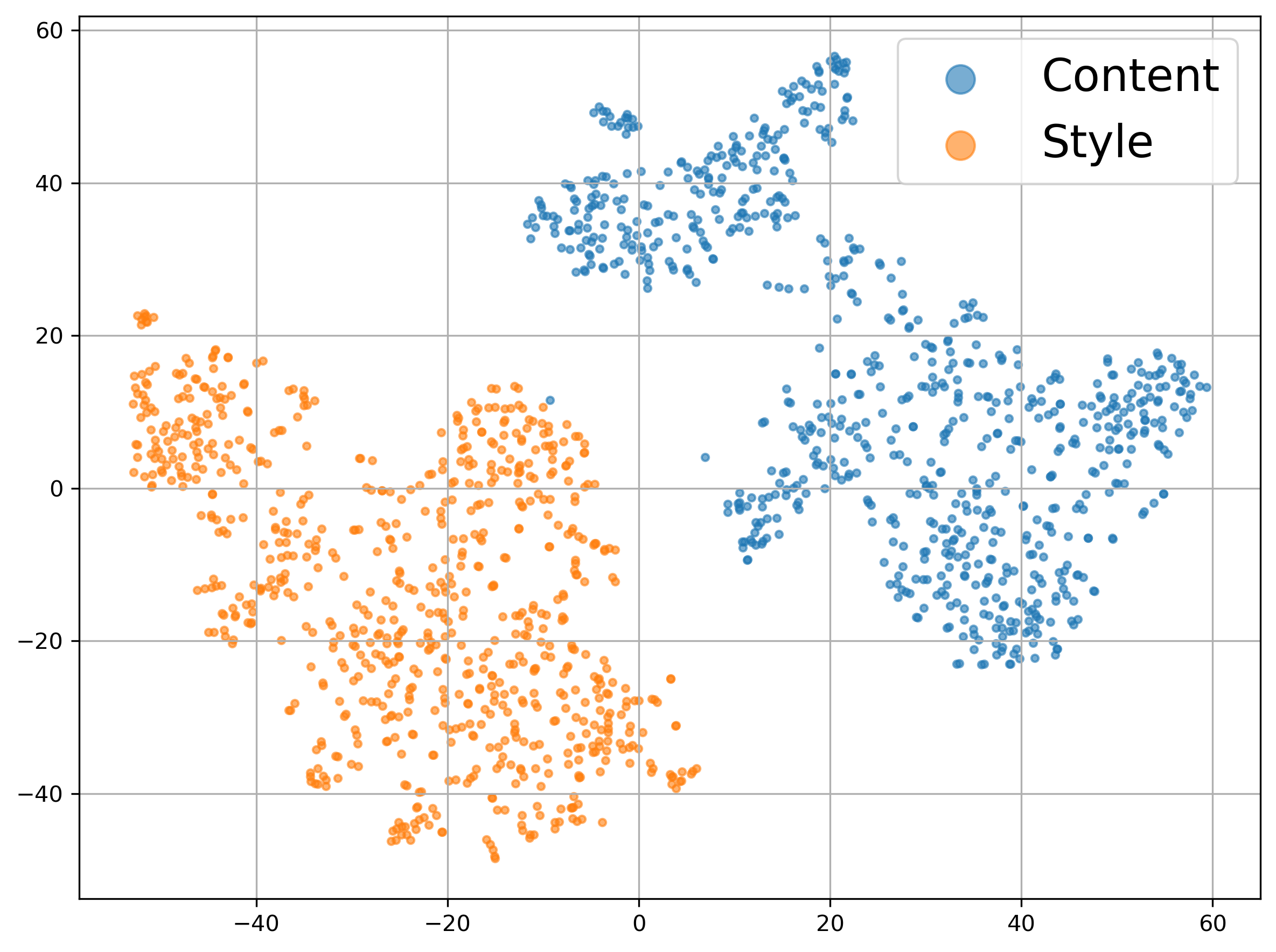}
    \caption{Content and style disentanglement shown by tSNE of WikiArt.
    }
    \label{fig:tsne_PH7R_3_d7P9_2}
\end{figure}

\section{More Analysis on mean content and style}
\label{sec:app:mean_content_style}
\begin{figure*}[htbp]
    \centering

    \includegraphics[width=1\linewidth, height=1\linewidth, keepaspectratio]{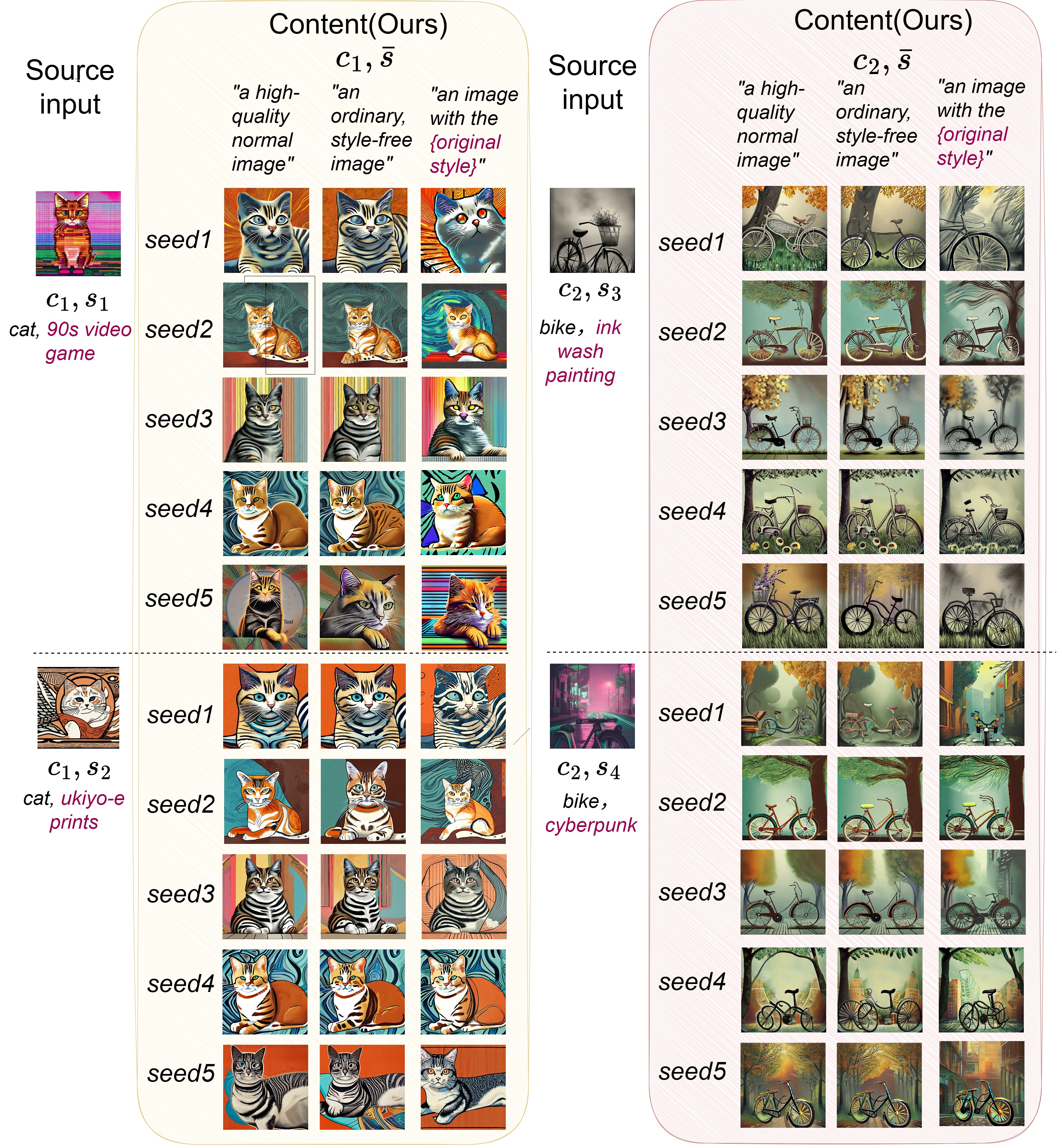}
    \caption{We use unCLIP to decode the same \textit{content embeddings} generated by our method to pixel space, using different prompts and initial noises (denoted by seed).
    }
    \label{fig:content_cls}
\end{figure*}

\begin{figure*}[htbp]
    \centering

    \includegraphics[width=1\linewidth, height=1\linewidth, keepaspectratio]{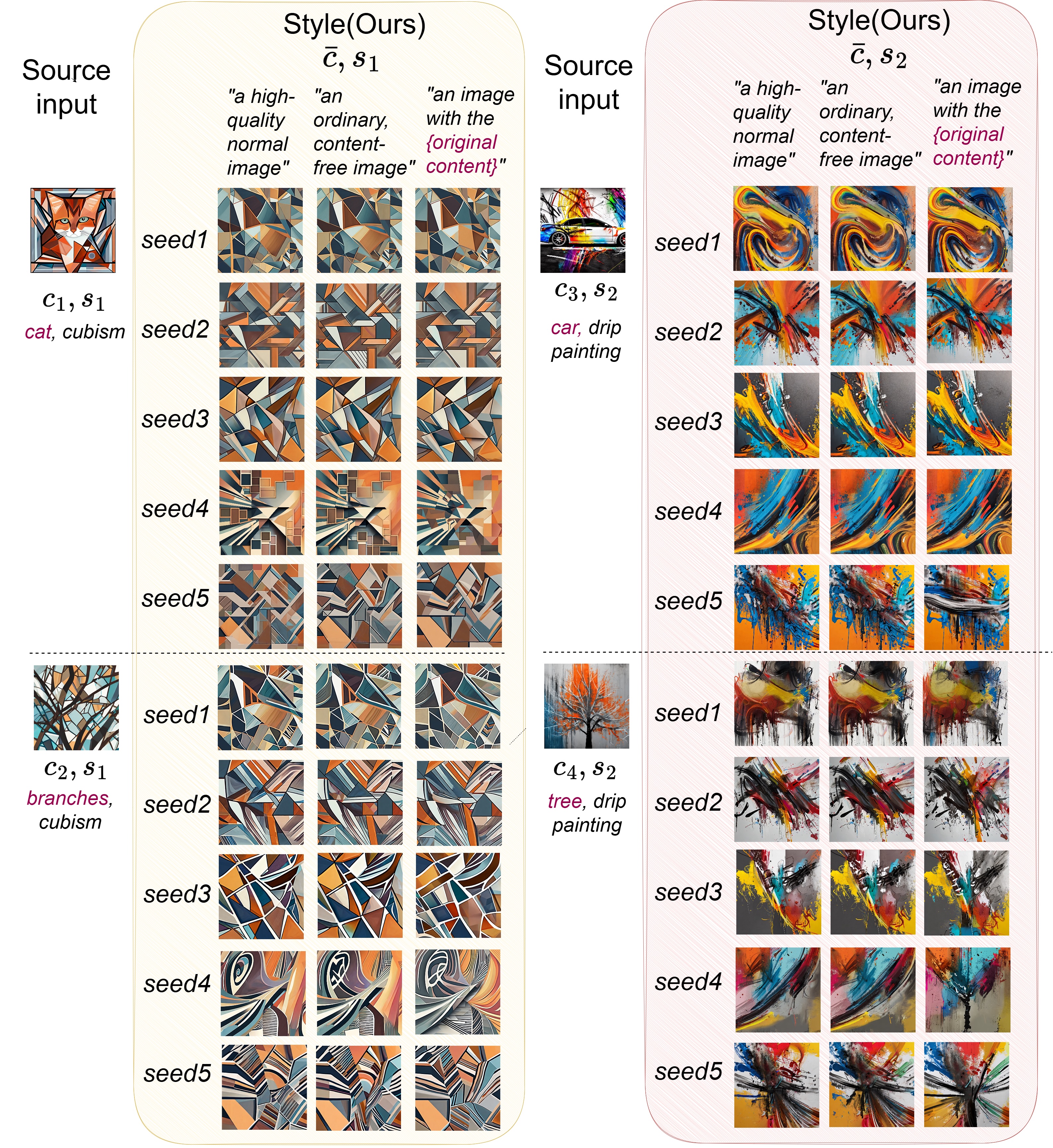}
    \caption{We use unCLIP to decode the same \textit{style embeddings} generated by our method to pixel space, using different prompts and initial noises (denoted by seed).
    }
    \label{fig:style_cls}
\end{figure*}

Our method produces disentangled representations $\bar{c}$ (content from style) and $\bar{s}$ (style from content), visualized in \cref{fig:content_cls,fig:style_cls}. Three key findings validate their independence and authenticity:

\begin{enumerate}
    \item \textbf{Content-Style Independence.} The disentangled embeddings exhibit no dependence on their original counterparts. For content embeddings $\bar{c}$, cat images rendered in diverse artistic styles all yield consistent $\bar{c}$ representations, as do bicycles (\cref{fig:content_cls}, right). Similarly, style embeddings $\bar{s}$ remain stable across different content inputs – for instance, the same style emerges whether applied to cats or branches (\cref{fig:style_cls}, left).
    
    \item \textbf{Intrinsic Signal Origin.} The $\bar{c}$ and $\bar{s}$ signals originate from our embeddings rather than unCLIP hallucinations. This is evidenced by two observations: (i) varying text prompts (different columns) produce negligible changes in $\bar{c}$ / $\bar{s}$, despite unCLIP's known prompt sensitivity; and (ii) our embedding signals consistently overpower prompt conditioning, maintaining semantic stability.
    
    \item \textbf{Initial Noise Invariance.} When testing different initial noise seeds in unCLIP, we observe minor variations in image details (\eg, object pose or texture) but noticeable consistency: $\bar{c}$ and $\bar{s}$ remain preserved across all noise configurations. This confirms their independence from generation artifacts.
\end{enumerate}

Collectively, these results demonstrate that $\bar{c}$ and $\bar{s}$ capture intrinsic content/style properties rather than inversion artifacts or model biases from unCLIP.

\section{Visualization of Unseen Styles and Contents}
\label{sec:app:unseen}
\paragraph{Unseen textual condition.}  
Our model is trained and evaluated solely on CLIP image embeddings, \textbf{without} using any text descriptions. Nevertheless, thanks to the multi-modal alignment in CLIP space, our model is capable of taking text as style and content references (\cref{fig:text_d7P9_3_PH7R_m1}) to generate meaningful results. 

\begin{figure*}[htbp]
    \centering
    \begin{subfigure}[t]{0.48\linewidth}
        \centering
        \includegraphics[width=.75\linewidth]{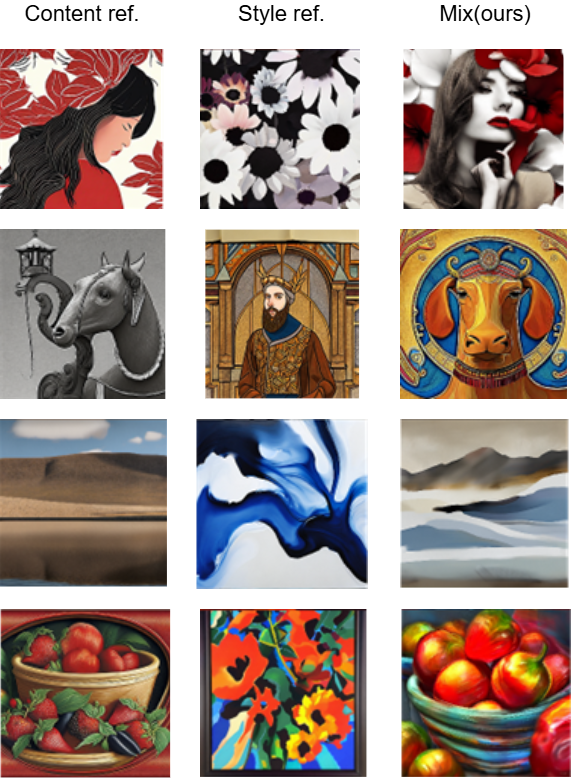}
        \caption{Mixing content and style references (unseen). The first and second columns show the content and style references, respectively; the third column shows the mixed results.}
        \label{fig:new_style_vis_mix}
    \end{subfigure}
    \hfill
    \begin{subfigure}[t]{0.48\linewidth}
        \centering
        \includegraphics[width=\linewidth]{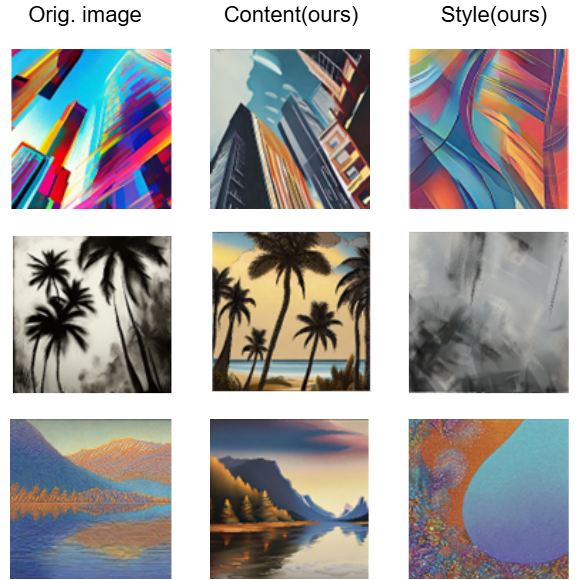}
        \caption{Disentanglement of content and style from unseen images. The first column shows the original image, followed by the extracted content and style.}
        \label{fig:new_style_vis_back}
    \end{subfigure}
    \caption{Visual results on unseen content and style inputs. Left: Mixing. Right: Disentanglement.}
\end{figure*}

\paragraph{Unseen constructed style and content.}
We curated an additional subset, similar to the main dataset, where both the style and content are never used during training or testing for our model. We visualize the corresponding forward and backward inference results. (see \cref{fig:new_style_vis_mix} and \cref{fig:new_style_vis_back})

\paragraph{Unseen real-world data from ImageNet and WikiArt.}  
Although not trained for the style and content retrieval task, our model yields competitive performance, reflecting strong representation quality. Our primary goal is to introduce a new possibility for semantic disentanglement with generative models. 
As shown in \cref{fig:tsne_PH7R_3_d7P9_2}, we achieve clear style-content separation on WikiArt (quantitative in \cref{tab:recall_appendix}). \cref{fig:ImageNet_PH7R_3_d7P9_2} further demonstrates successful forward and reverse inference on real photos (content ref.) and artworks (style ref.), showing generalization beyond synthetic data.

\section{Comparison with Conventional Discriminative Objectives}
\label{sec:app:contrastive}
In addition to CSD~\cite{somepalli2024CSD}, we train two models with \textit{Contrastive Loss}~\cite{chopra2005learning} and \textit{InfoNCE}~\cite{oord2018representation} on our dataset, using similar capacity, training settings, and evaluate on our test sets and real-world datasets using the same metrics. Except for a slight NMI on WikiArt, our method outperforms them across all settings (\cref{tab:recall_appendix}), confirming that our gains \textit{do not stem solely from the dataset}.
Importantly, our model learns a well-structured embedding space that enables style/content interpolation (\cref{fig:i_to_j} and \cref{fig:clipscore}) and avoids collapsing to mean interpretations, indicating strong generalization and balanced intra-/inter-variance. While simpler methods may work for single tasks, ours unifies merging and disentangling within a single framework.

\begin{table}[htbp]
    \centering
    \tiny
    \begin{tabular}{c||c|c|c|c|c|c}
        \hline
        \multirow{2}{*}{\textbf{Dataset}} & \multicolumn{3}{c}{\textbf{NMI Score}$\uparrow$} & \multicolumn{3}{c}{\textbf{FDR$\uparrow$}} \\ 
        & SCFlow & Contrastive~\cite{chopra2005learning} & InfoNCE~\cite{oord2018representation} & SCFlow & Contrastive & InfoNCE\\ 
        \midrule
        \rowcolor{gray!8}Our Styles & \textbf{0.8696}  & 0.2905 & \underline{0.5904} & \textbf{3.5184} & 0.1102 & \underline{0.3711} \\  
        WikiArt & 0.4010  & \underline{0.4194} & \textbf{0.4238} & \textbf{0.6474} & \underline{0.2923} & 0.2553  \\   
        \midrule
        \rowcolor{gray!8} Our Contents & \textbf{0.8356} & \underline{0.4598} & 0.2327 & 
        \textbf{2.1693} & \underline{0.1799} & 0.0598 \\  
        ImageNet  & \textbf{0.9172}  & 0.7737 & \underline{0.8194} &
        \textbf{1.4264} & \underline{0.3529} & 0.2056 \\  
        \bottomrule
    \end{tabular}
    
    \caption{Comparison to conventional discriminative approaches trained on our dataset.}
    \label{tab:recall_appendix}
\end{table}

\end{document}